\documentclass[12pt]{article}

\usepackage{amssymb,amsthm,amsmath,mathrsfs,bm,color,setspace}

\usepackage{lineno}
\usepackage{graphicx}
\usepackage{multirow}
\usepackage{booktabs}
\usepackage{epsfig}
\usepackage{natbib}
\usepackage{setspace}
\usepackage{threeparttable}

\usepackage{doi}
\usepackage{algorithm,algorithmic}

\newcommand{\defn}{\stackrel{\mbox{{\tiny def}}}{=}}
\newcommand{\trans}{^{\mbox{\tiny{T}}}}
\newtheorem{theo}{\bf Theorem}
\newtheorem{lemm}{\bf Lemma}[section]

\newtheorem{prop}{\bf Proposition}
\newtheorem{coro}{\bf Corollary}

\def\t{{\bf t}}
\def\x{{\bf x}}

\def\calD{\mathcal{D}}
\def\calN{\mathcal{N}}
\def\calS{\mathcal{S}}
\def\calG{\mathcal{G}}
\def\wh{\widehat}
\def\wt{\widetilde}

\def\beqr{\begin{eqnarray}}
\def\eeqr{\end{eqnarray}}
\def\beqrs{\begin{eqnarray*}}
\def\eeqrs{\end{eqnarray*}}

\def\ba{\bm{\alpha}}

\def\bSig{\mathbf{\Sigma}}

\def\mR{\mathbb{R}}

\newcommand{\strans}{^{*\mbox{\tiny{T}}}}

\newcommand{\md}{\mathrm{d}}

\newcommand\independent{\protect\mathpalette{\protect\independenT}{\perp}}
\def\independenT#1#2{\mathrel{\rlap{$#1#2$}\mkern2mu{#1#2}}}
\def \O{O}
\def \o{o}
\newcommand{\B}{\mathbf{B}}
\newcommand{\T}{\mathbf{T}}

\newcommand{\V}{\mathbf{V}}
\newcommand{\U}{\mathbf{U}}
\newcommand{\D}{\bm{D}}
\newcommand{\bv}{\mathbf{v}}
\newcommand{\bu}{\mathbf{u}}
\newcommand{\A}{\mathbf{A}}

\usepackage{url}
\DeclareFontFamily{U}{mathx}{\hyphenchar\font45}
\DeclareFontShape{U}{mathx}{m}{n}{
	<5> <6> <7> <8> <9> <10>
	<10.95> <12> <14.4> <17.28> <20.74> <24.88>
	mathx10
}{}
\DeclareSymbolFont{mathx}{U}{mathx}{m}{n}
\DeclareFontSubstitution{U}{mathx}{m}{n}
\DeclareMathAccent{\widecheck}{0}{mathx}{"71}
\DeclareMathAccent{\wideparen}{0}{mathx}{"75}

\newcommand{\bbeta}{\bm{\beta}}
\newcommand{\bgamma}{\bm{\gamma}}

\newcommand{\bhbeta}{\widehat{\bbeta}}

\newcommand{\argmin}{\mathop{\rm arg\min}}
\newcommand{\supp}{\mathrm{supp}}

\newcommand{\norm}[1]{\Vert#1\Vert}

\newcommand{\Inner}[2]{\left\langle #1, #2 \right\rangle}
\newcommand{\abs}[1]{\vert#1\vert}
\newcommand{\Abs}[1]{\left\vert#1\right\vert}

\newcommand{\W}{\mathbf{W}}

\DeclareMathOperator*{\pr}{\operatorname{pr}}
\DeclareMathOperator*{\expect}{\operatorname{E}}
\newcommand{\bzeros}{\bm{0}}

\newcommand{\bbetaT}{{\bbeta^\ast}}

\usepackage{caption}
\newcommand{\bmu}{\bm{\mu}}

\usepackage{tikz}
\tikzset{> = stealth,
    hidden/.style = {
        draw = black,
        shape = circle,
        inner sep = 1pt
    }
}

\usepackage[all,import]{xy}

\def\spacingset#1{\renewcommand{\baselinestretch}%
		{#1}\small\normalsize}

\usepackage{cleveref}
\crefname{theo}{Theorem}{Theorems}
\crefname{lemm}{Lemma}{Lemmas}
\crefname{coro}{Corollary}{Corollaries}
\crefname{prop}{Proposition}{Propositions}

\usepackage{enumitem}
\newlist{conditions}{enumerate}{10}
\setlist[conditions]{label*=(C\arabic*)}
\crefname{conditionsi}{condition}{conditions}
\Crefname{conditionsi}{Condition}{Conditions}
\crefrangeformat{conditionsi}{conditions~#3#1#4--#5#2#6}
\crefmultiformat{conditionsi}%
{conditions~#2#1#3}{ and~#2#1#3}{, #2#1#3}{ and~#2#1#3}
\crefrangemultiformat{conditionsi}%
{conditions~#3#1#4--#5#2#6}{ and~#3#1#4--#5#2#6)}{, #3#1#4--#5#2#6}{ and~#3#1#4--#5#2#6}

\Crefrangeformat{conditionsi}{Conditions~#3#1#4--#5#2#6}
\Crefmultiformat{conditionsi}%
{Conditions~#2#1#3}{ and~#2#1#3}{, #2#1#3}{ and~#2#1#3}
\Crefrangemultiformat{conditionsi}%
{Conditions~#3#1#4--#5#2#6}{ and~#3#1#4--#5#2#6}{, #3#1#4--#5#2#6}{ and~#3#1#4--#5#2#6}

\newcommand{\calL}{\mathcal{L}}

\usepackage{hyperref}
\usepackage{cleveref}
\usepackage{ulem}

\graphicspath{{figs/}}

\newcommand{\calI}{\mathcal{I}}
\newcommand{\calE}{\mathcal{E}}

\newcommand{\bp}{\mathbf{p}}

\usepackage{cleveref}
\crefname{theo}{Theorem}{Theorems}
\crefname{lemm}{Lemma}{Lemmas}
\crefname{coro}{Corollary}{Corollaries}
\crefname{prop}{Proposition}{Propositions}

\newlist{assumptions}{enumerate}{10}
\setlist[assumptions]{label*=(A\arabic*)}
\crefname{assumptionsi}{assumption}{assumptions}
\Crefname{assumptionsi}{Assumption}{Assumptions}
\crefrangeformat{assumptionsi}{assumptions~#3#1#4--#5#2#6}
\crefmultiformat{assumptionsi}%
{assumptions~#2#1#3}{ and~#2#1#3}{, #2#1#3}{ and~#2#1#3}
\crefrangemultiformat{assumptionsi}%
{assumptions~#3#1#4--#5#2#6}{ and~#3#1#4--#5#2#6}{, #3#1#4--#5#2#6}{ and~#3#1#4--#5#2#6}

\Crefrangeformat{assumptionsi}{Assumptions~#3#1#4--#5#2#6}
\Crefmultiformat{assumptionsi}%
{Assumptions~#2#1#3}{ and~#2#1#3}{, #2#1#3}{ and~#2#1#3}
\Crefrangemultiformat{assumptionsi}%
{Assumptions~#3#1#4--#5#2#6}{ and~#3#1#4--#5#2#6}{, #3#1#4--#5#2#6}{ and~#3#1#4--#5#2#6}

\DeclareFontFamily{U}{mathx}{\hyphenchar\font45}
\DeclareFontShape{U}{mathx}{m}{n}{
      <5> <6> <7> <8> <9> <10>
      <10.95> <12> <14.4> <17.28> <20.74> <24.88>
      mathx10
      }{}
\DeclareSymbolFont{mathx}{U}{mathx}{m}{n}
\DeclareFontSubstitution{U}{mathx}{m}{n}
\DeclareMathAccent{\widebar}{0}{mathx}{"73}

\usepackage{tikz}
\definecolor{myblack}{HTML}{000000}
\definecolor{myyellow}{HTML}{EFC000}
\definecolor{myred}{HTML}{D95319}
\definecolor{myblue}{HTML}{0073C2}
\definecolor{mygreen}{HTML}{7FCD61}

\DeclareRobustCommand{\uniformline}{%
\tikz[baseline=-0.6ex]{\draw[line width=0.65mm,myblack] (-0.35,0)--(0.35,0);}}

\DeclareRobustCommand{\laplacianline}{%
\tikz[baseline=-0.6ex]{\draw[line width=0.65mm,dashed,myyellow] (-0.35,0)--(0.35,0);}}

\DeclareRobustCommand{\logisticline}{%
\tikz[baseline=-0.6ex]{\draw[line width=0.65mm,dotted,myred] (-0.35,0)--(0.35,0);}}

\DeclareRobustCommand{\gaussianline}{%
\tikz[baseline=-0.6ex]{\draw[line width=0.65mm,dash pattern=on 4pt off 2pt,myblue] (-0.35,0)--(0.35,0);}}

\DeclareRobustCommand{\epanechnikovline}{%
\tikz[baseline=-0.6ex]{\draw[line width=0.65mm,dash pattern=on 4pt off 2pt on 1pt off 2pt,mygreen] (-0.35,0)--(0.35,0);}}
\usepackage{xr}

\usepackage{amsmath}
\usepackage{graphicx}
\usepackage{enumitem}
\usepackage{natbib}
\usepackage{url} 

\newcommand{\blind}{1}

\begin{document}

\def\spacingset#1{\renewcommand{\baselinestretch}%
{#1}\small\normalsize} \spacingset{1}


\if1\blind
{
  \title{\bf Efficient Distributed Learning over Decentralized Networks with Convoluted Support Vector Machine\footnote{The authors contributed equally to this work, and their names are listed alphabetically.  Liping Zhu (email:~\href{mailto:zhu.liping@ruc.edu.cn}{zhu.liping@ruc.edu.cn}) is the corresponding author.}}
  \author{\vspace{.2cm}Canyi Chen \\\vspace{.4cm}
    Department of Biostatistics, University of Michigan \\\vspace{.2cm}
    Nan Qiao \\\vspace{.4cm}
    School of Statistics, Renmin University of China \\\vspace{.2cm}
    Liping Zhu \\\vspace{.4cm}
    Institute of Statistics and Big Data, Renmin University of China
    }
  \maketitle
} \fi

\if0\blind
{
  \bigskip
  \bigskip
  \bigskip
  \begin{center}
    {\LARGE\bf  Efficient Distributed Learning over Decentralized Networks with Convoluted Support Vector Machine}
\end{center}
  \medskip
} \fi

\bigskip
\begin{abstract}
This paper addresses the problem of efficiently classifying high-dimensional data over decentralized networks. Penalized support vector machines (SVMs) are widely used for high-dimensional classification tasks. However, the {\it double} nonsmoothness of the objective function poses significant challenges in developing efficient decentralized learning methods. Many existing procedures suffer from slow, sublinear convergence rates. To overcome this limitation, we consider a convolution-based smoothing technique for the nonsmooth hinge loss function. The resulting loss function remains convex and smooth. We then develop an efficient generalized alternating direction method of multipliers (ADMM) algorithm for solving penalized SVM over decentralized networks.
Our theoretical contributions are twofold. First, we establish that our generalized ADMM algorithm achieves provable linear convergence with a simple implementation. Second, after a sufficient number of  ADMM iterations, the final sparse estimator attains near-optimal statistical convergence and accurately recovers the true support of the underlying parameters. 
Extensive numerical experiments on both simulated and real-world datasets validate our theoretical findings.
\end{abstract}

\noindent%
\vfill

\newpage
\spacingset{1.9} 
\addtolength{\textheight}{.5in}
\section{Introduction\label{section:introduction}}
Massive datasets, characterized by both large sample sizes and high-dimensional features, are increasingly prevalent across diverse fields. For example, the \cite{1000genomesprojectconsortium2015GlobalReferenceHuman} study amassed genomic data from 2,504 individuals spanning 26 populations, yielding approximately $12$ terabytes data. Often, such datasets are distributed across multiple locations. Fusing data together for centralized statistical analysis is somehow infeasible due to concerns over data privacy, memory and storage limitations, and bandwidth constraints. The absence of fusion centers has thus fueled interest in decentralized distributed learning---a paradigm that fully exploits distributed datasets by performing computations locally.  This methodology has found successful applications in fields such as personalized medicine, edge computing, 
smart utilities, and dimension reduction \citep{li2011PrincipalSupportVector}.  
A fundamental task in these applications is classification. 

Penalized support vector machines (SVMs) have been enduringly powerful tools for high-dimensional classification tasks, building on the seminal contributions of \cite{boser1992TrainingAlgorithmOptimal} and \cite{vapnik2000NatureStatisticalLearning}. The standard objective function for penalized SVMs combines the hinge loss with a penalty term. Commonly used  penalties include the $\ell_2$ penalty (as in the classical SVM), the $\ell_1$ penalty, SCAD \citep{park2012OraclePropertiesSCADpenalized, peng2016ErrorBoundL1norm}, and nuclear norm penalties \citep{xu2022DistributedEstimationSupporta}. Among these, the $\ell_1$ and elastic-net penalized SVMs, introduced by \cite{wang2006DoublyRegularizedSupport} and \cite{zhu20031normSupportVector}, are particularly notable for their ability to perform both classification and variable selection. Sparse decision rules enhance interpretability and stability---features highly desirable to practitioners such as physicians. Furthermore, \cite{peng2016ErrorBoundL1norm} rigorously analyzed the convergence behavior and error bounds of the $\ell_1$ penalized SVM, providing additional theoretical support for its utility.

Despite its advantages, the $\ell_1$ and elastic-net  penalized SVM presents considerable challenges due to its {\it doubly} nonsmooth objective function, which complicates the development of efficient algorithms, particularly in decentralized settings. Algorithms such as coordinate gradient descent \citep{friedman2010RegularizationPathsGeneralized} often face difficulties in achieving proper convergence owing to the nonsmooth nature of the hinge loss function \citep{luo1992LinearConvergenceDescent, tseng2001ConvergenceBlockCoordinate}. Similarly, classical approaches like interior-point methods and sequential minimal optimization \citep{platt1998SequentialMinimalOptimization} are generally ill-suited for high-dimensional problems due to scalability limitations. Extending these techniques to decentralized frameworks poses additional complexities.  Moreover, the fast {\it linear} convergence characteristic of most decentralized algorithms \citep{chang2014MultiAgentDistributedOptimization} typically relies on the smoothness of the objective function, such as that provided by least squares regression. These challenges lead to a fundamental research question: {\it Can we design a decentralized sparse classification method for $\ell_1$ and elastic-net penalized SVMs that achieves fast linear convergence?} 

We address this question affirmatively by proposing a novel method, called {\it decentralized penalized convoluted support vector machine (deCSVM)}. Our main contributions are outlined below: 
\begin{itemize} 
\item {\it A generalized alternating direction method of multipliers (ADMM) algorithm:} To mitigate the challenges posed by the nonsmooth hinge loss, we first consider a convolution-based smoothing technique \citep{wang2022DensityConvolutedSupportVector,chen2022CommunicationEfficientDistributedSupportVectorMachine}. This approach constructs a new loss function that is both smooth and convex, facilitating the development of efficient optimization algorithms for penalized SVMs. To derive sparse classification rules in decentralized networks, we present a generalized ADMM algorithm tailored to the penalized CSVM. This algorithm incorporates a quadratic majorization of the smoothed loss function, allowing all updates to have closed forms via efficient matrix-vector multiplications. 
\item {\it Linear optimization convergence rate:}  
In high-dimensional settings, the lack of strong convexity in the objective function often limits the convergence guarantees of classical ADMM. Existing theory typically establishes only sublinear convergence rates or convergence to a limiting point. By leveraging the strong convexity of the newly introduced smoothed loss function, however, we prove that the proposed algorithm enjoys fast {\it linear} convergence. This notable improvement ensures that the number of communication rounds required scales logarithmically with the desired accuracy, significantly enhancing computation and communication efficiency in decentralized settings.  
\item {\it Optimal statistical guarantees:} We establish optimal statistical guarantees on both estimation accuracy and support recovery for our decentralized estimation process. After a sufficient number of ADMM iterations, the proposed estimator attains the optimal statistical convergence rate, matching that of an equivalent in-memory penalized SVMs where all data are pooled. Moreover, we prove that the method achieves exact support recovery under the ``beta-min'' condition, providing rigorous theoretical assurance of its effectiveness.  To the best of our knowledge, this is perhaps the first result for support recovery guarantees of elastic-net penalized SVM. 
\end{itemize}

\subsection{Related Works}
{
Numerous methods have been proposed in the distributed learning literature, broadly categorized into two main frameworks. {\it Centralized distributed learning:} A central node coordinates the learning process, as explored in \cite{hector2020DoublyDistributedSupervised, hector2021DistributedIntegratedMethod, tang2020DistributedSimultaneousInference, zhou2024DistributedEmpiricalLikelihooda, chen2023DistributedDecodingHeterogeneous}. However, these methods face limitations, including bandwidth constraints at the central node and vulnerability to system-wide failure in the event of node disruption. {\it Decentralized distributed learning:} When a central node is unavailable or impractical, decentralized learning offers a robust alternative. For instance, \cite{yadav2007distributed} employed an averaging consensus protocol leveraging the spectral properties of the Laplacian matrix, although it often produced undesired dense estimates. \cite{chang2014MultiAgentDistributedOptimization} introduced inexact consensus ADMM with linear convergence guarantees for smooth, strongly convex objectives, but the method is unsuitable for high-dimensional settings. Further work by \cite{shi2015ProximalGradientAlgorithm} and \cite{li2019DecentralizedProximalGradientMethod} developed decentralized (proximal) gradient descent algorithms that eschew the use of Lagrangian duals. Subgradient methods have also been applied to quantile regression (QR) in \cite{wang2018DistributedQuantileRegression, wang2019DistributedOnlineQuantile, zhang2019DistributedDiscreteTimeOptimization}, though they generally yield dense estimates and prioritize optimization convergence over statistical guarantees. 

Recent developments in distributed SVM methods have sought to capitalize on parallel computing techniques. \cite{lian2018DivideandConquerDebiasedNorm} proposed a divide-and-conquer strategy for $\ell_1$ penalized SVMs in centralized settings, requiring each local node to store at least $N^{2/3}\log(p)$ observations (where $N$ is the sample size and $p$ the covariate dimension) to achieve optimal statistical accuracy. \cite{wang2019DistributedInferenceLinear} addressed this limitation with a linear estimate of the standard SVM (LESVM) using kernel smoothing, but this approach still relies on a central node for storing a subset of the data and broadcasting updated estimates. Other distributed SVM approaches utilize the approximate Newton methods; see \cite{zhou2022CommunicationEfficientDistributedLearning} for $\ell_1$ penalized SVM and \cite{xu2022DistributedEstimationSupporta} for nuclear norm penalized SVM. However, these methods struggle with the nonsmooth hinge loss, necessitating substantial data storage at each local node, impractical in memory-constrained environments.

Several recent works on quantile regression (QR) are also relevant. In decentralized settings, \cite{wang2018DistributedQuantileRegression, wang2019DistributedInferenceLinear, zhang2019DistributedDiscreteTimeOptimization} proposed decentralized sub-gradient algorithms for QR. However, sub-gradient-based algorithms typically exhibit slow convergence rates. 
To address this, \cite{liu2022FastRobustSparsity} recast the QR loss as a least squares loss, achieving faster convergence under $p \leq n$ (where $n$ is the local sample size). Related work by \cite{lu2023SparseLowRankMatrix} focused on regularized matrix QR problems with extension by \cite{chen2022RobustFastLowRank} to decentralized settings. However, these approaches, including \cite{chen2024DecentralizedDistributedEstimation}, frequently involve nested iterations, leading to significant communication costs.

To handle nonsmooth loss functions, smoothing techniques have gained attention.  \cite{horowitz1998BootstrapMethodsMedian} introduced a smoothing method for the indicator function in QR loss, further refined by \cite{galvao2016SmoothedQuantileRegression} and \cite{chen2019QuantileRegressionMemory}. \cite{wang2019DistributedInferenceLinear} adapted these ideas for standard SVMs, though at the expense of convexity. More recently, inspired by \cite{rubinstein1983SmoothedFunctionalsStochastic}, \cite{fernandes2021SmoothingQuantileRegressions} developed a method preserving both convexity and smoothness, which was extended to sparse penalized QR by \cite{tan2021HighDimensionalQuantileRegression} with demonstrated statistical properties. For SVMs, \cite{wang2022DensityConvolutedSupportVector} investigated elastic-net penalized SVMs in high-dimensional settings, while \cite{chen2022CommunicationEfficientDistributedSupportVectorMachine} examined nuclear norm penalized SVMs. However, extending these methods to decentralized settings remains an open challenge.

}





\subsection{Paper Organization and Notations}
The paper is organized as follows. In \Cref{section:methodology}, we present the decentralized penalized SVM framework for networked settings. \Cref{section:theory} outlines the theoretical guarantees of the proposed method. 
Extensive simulation studies are detailed in \Cref{section:numerical_studies}, while \Cref{section:application} demonstrates the practical utility of our approach using the UCI Communities and Crime dataset.   \Cref{section:conclusion} concludes the paper with discussions on potential future directions. Technical details and additional simulations are provided in the Supplementary Material.

{
We use the following notations. $C, C_0, C_1, \ldots, c, c_0, c_1,\ldots$ represent generic constants that may vary across instances. 
Standard asymptotic notation is also used.  
Given two sequences
$\{a_n\}$ and $\{b_n\}$, we write $a_n = \O(b_n)$ or $a_n\lesssim b_n$ if there exists a constant
$C<\infty$ such that $\abs{a_n}\leq Cb_n$,
and $a_n = \o(b_n)$ if $a_n/b_n\to 0$.
For two sets of random variables
$\{X_n\}$ and $\{Y_n\}$, we write $X_n = \O_p(Y_n)$ if for any $\epsilon>0$,
there exists a finite $M>0$ and a finite $n_0>0$ such that
$\pr(\abs{X_n/Y_n}>M)<\epsilon$ for any $n>n_0$. 
For a vector $\bv = (v_1,\ldots,v_p)\trans$, we define $\abs{\bv}_1 \defn
\sum_{j=1}^p\abs{v_j}$, $\abs{\bv}_2 \defn
(\sum_{j=1}^pv_j^2)^{1/2}$ and $\abs{\bv}_\infty \defn
\max_{j}\abs{v_j}$. 
 For a matrix $\A=(a_{i,j})\in\mR^{n_1\times n_2}$, denote by $\Lambda_{\max}(\A)$ and $\Lambda_{\min}(\A)$ the largest and smallest singular values of $\A$.
 Let $\norm{\A} = \Lambda_{\max}(\A)$ and $\norm{\A}_\infty = \max_{i}\sum_{j = 1}^{n_2}\abs{A_{i,j}}$ 
  be the spectral norm, and the infinity norm, respectively. 
  For sets $\calS_1\subseteq \{1, \ldots, n_1\}$ and $\calS_2\subseteq \{1, \ldots, n_2\}$, we use $\A_{\calS_1\calS_2} = (A_{i,j})_{i\in\calS_1, j\in\calS_2}$ denote the submatrix indexed by $\calS_1$ and $\calS_2$.
}



\section{Methodology}\label{section:methodology}

\subsection{Model Setup}
In this subsection, we outline the model setup. We consider a decentralized network consisting of \( m \) local nodes capable of performing computations and exchanging messages with their neighbors. These nodes are geographically distributed and may represent diverse entities such as 
routers monitoring Internet traffic, research laboratories, hospitals participating in large cohort studies, or mobile devices.  
The network is modeled as an undirected graph \( \mathcal{G} = (\mathcal{N}, \mathcal{E}) \), where \( |\mathcal{N}| = m \). The vertex set \( \mathcal{N} = \{1, \ldots, m\} \) corresponds to the local nodes, and the edge set \( \mathcal{E} \) comprises pairs of nodes capable of direct communication. A node \( \ell \) is restricted to communicating with its one-hop neighbors, denoted as \( \mathcal{N}(\ell) \) to minimize communication costs.  In symbols, the set of neighboring nodes for node \(\ell\) is denoted as \(\calN(\ell) \defn \{ k \colon (\ell, k) \in \calE \}\). 
The network's connectivity structure is encoded in the adjacency matrix \( \mathbf{W} \in \{0, 1\}^{m \times m} \). Specifically, \( W_{\ell k} = W_{k \ell} = 1 \) if and only if nodes \( \ell \) and \( k \) are connected by a direct link. By assumption, the network excludes self-loops, ensuring that all diagonal entries of \( \mathbf{W} \) are zero. Furthermore, the network is assumed to be connected, implying the existence of a (potentially multi-hop) path between any pair of nodes.


For \(\ell = 1, \ldots, m\), let \(\calI_\ell\) represent the index set of the data available at node \(\ell\). The sets \(\calI_\ell\) are disjoint and satisfy \(\bigcup_{\ell = 1}^m \calI_\ell = \{1, \ldots, N\}\), with \(|\calI_\ell | = n_\ell\). 
The data subset available at node \(\ell\) is denoted as \(\calD_\ell \defn \{ (\x_i, Y_i) \colon i \in \calI_\ell \}\), 
where \(\x_i = (X_{i1}, \ldots, X_{ip})\trans \in \mR^{p}\)  represents the \(p\)-dimensional covariate vector with \(X_{i1} \equiv 1\), 
and \(Y_i \in \{1, -1\}\) denotes the class label.  The complete dataset across all nodes is denoted by \( \calD \defn \bigcup_{\ell=1}^m \calD_\ell \). For simplicity, we assume that the data are evenly distributed, meaning \( n_\ell = n \) for all \( \ell \). 
 Extending the methodology to handle uneven local sample sizes is straightforward and requires minimal adjustment.

With these computing nodes in place, we aim to collaboratively and distributedly solve the following network-wide elastic-net penalized SVM problem \citep{zou2005RegularizationVariableSelection}:  
\beqr\label{loss:network}  
\min_{\bbeta \in \mR^{p}} \calL(\bbeta; \calD) = \min_{\bbeta \in \mR^{p}} \frac{1}{m} \sum_{\ell=1}^m \calL(\bbeta; \calD_\ell),  
\eeqr  
where \( \bbeta = (\beta_1, \ldots, \beta_p)\trans \) represents the separating hyperplane in \( \mR^p \), and \( \calL(\bbeta; \calD_\ell) \defn (n)^{-1} \sum_{i \in \calI_\ell} L(Y_i \x_i\trans \bbeta) + {\lambda_0} \abs{\bbeta}_2^2/2 + \lambda \abs{\bbeta}_1 \) denotes the elastic-net penalized SVM objective specific to node \( \ell \). The function \( L(u) = (1 - u)_+ = \max(1 - u, 0) \) is the hinge loss function. The objective \( \calL(\bbeta; \calD_\ell) \) reduces to the \( \ell_1 \)-penalized SVM as a special case when \( \lambda_0 = 0 \).  
The penalty parameters \( \lambda_0 \) and \( \lambda \) are global across the network \( \calG \). The parameter \( \lambda \) governs the sparsity of the estimated coefficients \( \bbeta \). For simplicity of presentation, we include the intercept \( \beta_1 \) in the penalty term, although it is typically excluded. The case of an unpenalized intercept can be easily handled by employing a weighted penalty.

In the decentralized network, a central coordinator node is unavailable to maintain high-dimensional global model parameters \( \bbeta \) during iterative optimization \citep{nedic2009DistributedSubgradientMethods, cao2013OverviewRecentProgress, sayed2014AdaptationLearningOptimization}. To address this limitation, we reformulate problem \eqref{loss:network} into an equivalent consensus form as follows:  
\beqr\label{loss:consensus}  
\min\limits_{\{\bbeta^{(\ell)}\}_{\ell=1}^m} \frac{1}{m} \sum_{\ell=1}^m \calL(\bbeta^{(\ell)}; \calD_\ell)  
\quad \mathrm{s.t.} \quad \bbeta^{(\ell)} = \bbeta^{(k)}, \ \forall (\ell, k) \in \calE,  
\eeqr  
where \( \bbeta^{(\ell)} \) represents a local copy of the parameters maintained at node \( \ell \). In this formulation, the constraints enforce equality between the local parameters of each node and those of its neighbors, giving rise to the term ``consensus''. 
Since the graph \( \calG \) is assumed to be connected, the solution to problem \eqref{loss:consensus} is equivalent to that of problem \eqref{loss:network}.   
 
The pursuit of a solution to the decentralized penalized SVM problem \eqref{loss:consensus} faces two significant challenges. First, the hinge loss function is nonsmooth, which complicates the application of gradient-based or Newton-based optimization algorithms. Second, the inclusion of the elastic-net penalty renders problem \eqref{loss:consensus} devoid of closed-form solutions in general.  
To address these challenges concurrently, we propose a decentralized convoluted SVM, detailed in the following sections. We begin by constructing a convolution-based smoothing procedure for the hinge loss function, resulting in a new loss function that is both convex and smooth. Subsequently, we introduce a generalized ADMM algorithm to solve the decentralized elastic-net penalized convoluted SVM efficiently. This generalized ADMM leverages a local majorization approximation, enabling each local update to have a closed-form solution, thereby significantly reducing computational costs.  
This efficiency is particularly critical in scenarios such as wireless sensing networks, where local nodes are often inexpensive, battery-powered, and computationally constrained.  
As demonstrated in \Cref{section:theory}, the convexity and smoothness of the smoothed loss ensure that the approximation does not impair convergence. In particular, we will show that the proposed optimization algorithm converges {\it linearly}.

\subsection{A Convolution-type Smoothing Procedure}
In this subsection, we consider a convolution-type smoothing procedure for the hinge loss function \citep{wang2022DensityConvolutedSupportVector,chen2022CommunicationEfficientDistributedSupportVectorMachine}. The resulting smoothed loss function is both convex and smooth, offering significant advantages for both theoretical analysis and computational efficiency.  

The key idea is to replace the empirical probability measure with a kernel-smoothed probability measure to construct the empirical risk in \eqref{loss:network}.  

To begin, define a new random variable \( U = Y\x\trans\bbeta \) with {\( F(\cdot; \bbeta) \)} denoting its cumulative distribution function (CDF). The population risk in \eqref{loss:network} can then be expressed as  
\beqrs  
\expect\{L(Y\x\trans\bbeta)\} = \int_{-\infty}^\infty L(t) \md F(t; \bbeta).  
\eeqrs  
If the CDF \( F(\cdot; \bbeta) \) is sufficiently smooth, one can expect that \( \expect\{L(Y\x\trans\bbeta)\} \) is at least twice differentiable and convex.

For each \( \bbeta \in \mR^p \), let  $\wh{F}(t; \bbeta) = \frac{1}{N} \sum_{i=1}^N I(Y_i\x_i\trans\bbeta \le t)  $ 
be the empirical probability measure based on an independently and identically distributed realization of \( U \), where \( I(\cdot) \) is the indicator function. The empirical risk in \eqref{loss:network} can then be written as  
\beqrs  
\int_{-\infty}^\infty L(t) \md \wh{F}(t; \bbeta).  
\eeqrs  
However, the empirical probability measure \( \wh{F}(\cdot; \bbeta) \) is discontinuous, causing the resulting loss function in \eqref{loss:network} to retain the nonsmoothness of the hinge loss \( L(\cdot) \). This limitation motivates the use of a smooth probability measure \( \wt{F}(\cdot; \bbeta) \) as an alternative estimate for the population CDF.

Specifically, we propose using the kernel density estimate for the population CDF:  
\beqrs  
\wt{F}(t; \bbeta) = \int_{-\infty}^t \frac{1}{Nh} \sum_{i=1}^N K\left( \frac{u - Y_i\x_i\trans\bbeta}{h} \right) \md u,  
\eeqrs  
where \( K: \mR \rightarrow [0, \infty) \) is a smooth kernel function satisfying \( K(-u) = K(u) \) for all \( u \in \mR \), \( \int_{-\infty}^\infty K(u) \md u = 1 \), and \( \int_{-\infty}^\infty |u| K(u) \md u < \infty \). The parameter \( h > 0 \) is a bandwidth.  

Replacing \( \wh{F} \) with \( \wt{F} \) results in a new loss function:  
\beqrs  
\int_{-\infty}^\infty L(t) \md \wt{F}(t; \bbeta)  =  \frac{1}{N} \sum_{i=1}^N \int_{-\infty}^\infty L(t) \frac{1}{h} K\left(\frac{t - Y_i\x_i\trans\bbeta}{h}\right) \md t \defn \frac{1}{N} \sum_{i=1}^N L_h(Y_i\x_i\trans\bbeta),  
\eeqrs  
where 
\( L_h(t) = \int_{-\infty}^\infty L(u) h^{-1} K\{(u - t)/h\} \md u \). By construction, the new smoothed hinge loss function \( L_h(\cdot) \) is both convex and smooth. Furthermore, it satisfies the relationship \( L_h = L * K_h \), where \( K_h(u) = h^{-1} K(u/h) \), and the operator \( * \) denotes convolution.  

We examine the proposed loss function in conjunction with several commonly used kernel functions. Additional examples are presented in {Section A} of the Supplementary Material.
\begin{enumerate}
    \item (Laplacian kernel) For the Laplacian kernel $K(u) = \exp(-\abs{u})/2$, the resulting smoothed hinge loss is expressed as
     $\calL_h^L(v) = [1 + h/2\exp\{(v - 1)/h\} - v]I(v<1) + h/2\exp\{(1-v)/h\}I(v\geq 1)$. 
    \item (Logistic kernel) For the logistic kernel $K(u) = \exp(-u)/\{1 + \exp(-u)\}^2$, the corresponding smoothed hinge loss is given by
     $\calL_h^{Logit}(v) = -v + h\log\{\exp(1/h) + \exp(v/h)\}$. 
    \item (Gaussian kernel) For the Gaussian kernel $K(u)=(2\pi)^{-1/2}\exp(-u^2/2)$, 
    the smoothed hinge loss becomes $\calL_h^G(v)=(1-v)\Phi\left\{(1-v)/h\right\}+h(2\pi)^{-1/2}\exp\left\{-(1-v)^2/(2h^2)\right\}$,
where $\Phi(\cdot)$ denotes the cumulative distribution function of the standard normal distribution.   
\end{enumerate}
The proposed classifier offers computational advantages by addressing the nonsmooth nature of the original hinge loss. This smoothing approach is rooted in the mollification technique introduced by \citet{friedrichs1944IdentityWeakStrong} and extensively studied in the optimization literature \citep[e.g.,][]{rubinstein1983SmoothedFunctionalsStochastic}. Recently, such methodologies have been increasingly adopted in statistical contexts \citep{he2021SmoothedQuantileRegression,fernandes2021SmoothingQuantileRegressions,tan2021HighDimensionalQuantileRegression,wang2022DensityConvolutedSupportVector}. Prominent examples include the Huberized smoothing approximation of hinge loss by \citet{rosset2007PiecewiseLinearRegularized} and \citet{wang2008HybridHuberizedSupport}, which facilitated the computation of elastic-net penalized SVMs through smoothed optimization techniques.
Our proposed loss function aligns with this class of convolution-based smoothing methods by utilizing $K$ as a generalized kernel. The theoretical framework developed in this study specifically accommodates the proposed loss function and ensures its applicability within this context.

By replacing the hinge loss function in problem \eqref{loss:network} with the proposed smoothed hinge loss, we formulate the following optimization problem:
\beqr\label{optimization:elastic_net}
\wh\bbeta = \argmin_{\bbeta \in \mR^{p}} \frac{1}{N} \sum_{i = 1}^N L_h(Y_i\x_i\trans\bbeta) + {\lambda_0/2\abs{\bbeta}_2^2 + \lambda\abs{\bbeta}_1}.
\eeqr
The corresponding network-wide consensus-aware formulation is given by
\beqr\label{problem:consensus} 
& \min\limits_{\B}  & \frac{1}{m}\sum_{\ell = 1}^m \left \{\frac{1}{n}\sum_{i \in \calI_\ell}L_h(Y_i\x_i\trans\bbeta^{(\ell)}) +  \lambda_0/2 \abs{\bbeta^{(\ell)}}_2^2 + \lambda \abs{\bbeta^{(\ell)}}_1\right\}\\
\notag & \mathrm{s.t.} &\  \bbeta^{(\ell)} = \bbeta^{(k)}\enskip \forall\ (\ell,k)\in\calE,
\eeqr
where $\B\defn(\bbeta^{(1)}, \ldots, \bbeta^{(m)})\trans\in\mR^{m\times p}$ is the concatenated version of the parameter vectors $\bbeta^{(\ell)}$.

{
The parameter $h$ serves as the bandwidth for the smoothing kernel and should be chosen small to ensure that the convoluted SVM closely approximates the original SVM. The specific choice of $h$ that ensures optimal statistical performance will be given in \Cref{theorem:convergence_rate} of  \Cref{section:theory}. 
}



\subsection{Generalized ADMM}
The simultaneous smoothness and convexity of the proposed loss function facilitate the development of an efficient decentralized algorithm to solve \eqref{problem:consensus}. The algorithm is provably guaranteed to converge linearly to a minimizer of \eqref{problem:consensus} while maintaining a straightforward implementation.

By introducing auxiliary variables $\T\defn (\t^{(\ell k)})_{\ell = 1, k}^{m, \calN(\ell)}$, problem \eqref{problem:consensus} is equivalent to
\beqr\label{problem:auxiliary}
& \min\limits_{\B, \T}  & \frac{1}{m}\sum_{\ell = 1}^m \left \{\frac{1}{n}\sum_{i \in \calI_\ell}L_h(Y_i\x_i\trans\bbeta^{(\ell)}) +  \lambda_0/2 \abs{\bbeta^{(\ell)}}_2^2 + \lambda \abs{\bbeta^{(\ell)}}_1\right\}\\
\notag & \mathrm{s.t.} &\  \bbeta^{(\ell)} = \bbeta^{(k)} = \t^{(\ell k)} \enskip \forall\ (\ell,k)\in\calE.
\eeqr
The linear constraint structure in the above formulation naturally suggests leveraging the  ADMM framework. Based on the classic ADMM framework \citep{boyd2010DistributedOptimizationStatistical}, we construct the augmented Lagrangian with a penalty parameter $\tau$ as follows:
\begin{equation}
	\begin{aligned}
L_{\tau}(\B, \T, \U, \V) &=  
\sum_{\ell = 1}^m \left\{\frac{1}{n}\sum_{i \in \calI_\ell}L_h(Y_i\x_i\trans\bbeta^{(\ell)})   + \lambda_0/2 \abs{\bbeta^{(\ell)}}_2^2 + \lambda \abs{\bbeta^{(\ell)}}_1\right\}   \\
	\label{eq:Lagrangian}	  &  + \sum_{\ell = 1}^m\sum_{k \in\calN(\ell)}\bigg(\langle\bu^{(\ell k)},   \bbeta^{(\ell)} - \t^{(\ell k)}\rangle +  \langle\bv^{(\ell k)},    \bbeta^{(k)} - \t^{(\ell k)}\rangle  \\
		  &   + \frac{\tau}{2}\abs{\bbeta^{(\ell)} - \t^{(\ell k)}}_2^2   + \frac{\tau}{2}\abs{\bbeta^{(k)} - \t^{(\ell k)}}_2^2 \bigg), 
	\end{aligned}
\end{equation}
where $\U\defn (\bu^{(\ell k)})_{\ell = 1, k}^{m, \calN(\ell)}$ and $\V\defn (\bv^{(\ell k)})_{\ell  = 1, k}^{m, \calN(\ell)}$ are introduced dual variables. Define variable $\bp_{t}^{(\ell)}\defn\sum_{k\in\calN(\ell)}(\bu_{t}^{(\ell k)} + \bv_{t}^{(\ell k)})$ with $\bp_{0}^{(\ell)} = \bm{0}$. To find the solution for the above augmented Lagrangian, we can recursively perform  \eqref{ADMM:beta}--\eqref{update:Pbeta}:
\begin{subequations}
	\begin{align}
 \bbeta_{t + 1}^{(\ell)} & =  \argmin_{\bbeta^{(\ell)}}  
\frac{1}{n}\sum_{i \in \calI_\ell}L_h(Y_i\x_i\trans\bbeta^{(\ell)})   + \lambda_0/2 \abs{\bbeta^{(\ell)}}_2^2 + \lambda \abs{\bbeta^{(\ell)}}_1  + \Inner{\bp_{t }^{(\ell)}}{\bbeta^{(\ell)}}  
		\label{ADMM:beta}  \\
		 \nonumber & +   \tau\sum_{k \in \calN(\ell)}\Abs{\bbeta^{(\ell)} - (\bbeta_{t}^{(\ell)} + \bbeta_{t}^{(k)})/2}_2^2,\\
		\label{update:Pbeta} \bp_{t + 1}^{(\ell)} & =  \bp_{t}^{(\ell)} + \tau\sum_{k\in\calN(\ell)} (\bbeta_{t+1}^{(\ell)} - \bbeta_{t+1}^{(k)}).
	\end{align}
\end{subequations}
Unfortunately, obtaining {$\bbeta_{t + 1}^{(\ell)}$} from problem \eqref{ADMM:beta} generally lacks a closed-form solution, necessitating multiple optimization iterations to approximate the solution. This process can lead to substantial computational overhead, particularly as these costs accumulate across ADMM iterations. To address this, we propose leveraging the Majorize-Minimization technique to perform an inexact update for \eqref{ADMM:beta}, as follows:  
	\begin{align}
\nonumber {\bbeta_{t + 1}^{(\ell)}} & \approx  \argmin_{\bbeta^{(\ell)}}  \frac{1}{n}\sum_{i \in \calI_\ell} L_h^\prime(Y_i\x_i\trans\bbeta^{(\ell)}_t)Y_i\x_i\trans(\bbeta^{(\ell)} - \bbeta^{(\ell)}_t)  + \frac{1}{2n}\sum_{i \in \calI_\ell} L_h^{\prime\prime}(Y_i\x_i\trans\bbeta^{(\ell)}_t)\x_i\x_i\trans\abs{\bbeta^{(\ell)} - \bbeta^{(\ell)}_t}_2^2\\
 \notag &  + \lambda_0/2 \abs{\bbeta^{(\ell)}}_2^2 + \lambda \abs{\bbeta^{(\ell)}}_1  + \Inner{\bp_{t}^{(\ell)}}{\bbeta^{(\ell)}}    +   \tau\sum_{k \in \calN(\ell)}\Abs{\bbeta^{(\ell)} - (\bbeta_{t}^{(\ell)} + \bbeta_{t}^{(k)})/2}_2^2\\
   \notag & \approx  \argmin_{\bbeta^{(\ell)}}  \frac{1}{n}\sum_{i \in \calI_\ell} L_h^\prime(Y_i\x_i\trans\bbeta^{(\ell)}_t)Y_i\x_i \trans(\bbeta^{(\ell)} - \bbeta^{(\ell)}_t) + {\rho_\ell/2  } \abs{\bbeta^{(\ell)} - \bbeta^{(\ell)}_t}_2^2\\
\label{problem:approximation} &  + \lambda_0/2 \abs{\bbeta^{(\ell)}}_2^2 + \lambda \abs{\bbeta^{(\ell)}}_1  + \Inner{\bp_{t}^{(\ell)}}{\bbeta^{(\ell)}}   +   \tau\sum_{k \in \calN(\ell)}\Abs{\bbeta^{(\ell)} - (\bbeta_{t}^{(\ell)} + \bbeta_{t}^{(k)})/2}_2^2.
	\end{align}
Here, $\rho_\ell > 0$ is a penalty parameter chosen such that $\rho_\ell \geq c_h \Lambda_{\max}\left(\frac{1}{n}\sum_{i \in \calI_\ell} \x_i\x_i\trans\right)$, where $c_h > 0$ denotes the Lipschitz constant of $L_h^\prime(\cdot)$, the first derivative of $L_h(\cdot)$.

It is easy to show that problem \eqref{problem:approximation} has a closed-form solution by invoking the proximal operator of the $\ell_1$ norm $\abs{\cdot}_1$,
\begin{equation}
{\bbeta_{t + 1}^{(\ell)}}   = \calS_{\lambda \omega_\ell}\Bigg[\omega_\ell\bigg\{\rho_\ell\bbeta_{t}^{(\ell)} - \frac{1}{n}\sum_{i \in \calI_\ell} L_h^\prime(Y_i\x_i\trans\bbeta^{(\ell)}_t)Y_i\x_i     - \bp_{t}^{(\ell)}
  + \tau \sum_{k \in\calN(\ell)}(\bbeta_{t}^{(\ell)} + \bbeta_{t}^{(k)})\bigg\}\Bigg], \label{update:beta}\tag{\ref{ADMM:beta}$^\prime$}
\end{equation}
where $\omega_\ell = 1/(2\tau \abs{\calN(\ell)} + \rho_\ell + \lambda_0)$, and $\calS_t(\bv)\defn (\bv - t \bm{1}_p)_{+} - (-\bv - t\bm{1}_p)_{+}$ denotes the coordinate-wise soft-thresholding operator.  
In summary, updates \eqref{update:beta} and \eqref{update:Pbeta} are performed locally at each node and constitute the generalized ADMM algorithm, as outlined in \Cref{alg:decsvm}. {Detailed derivations for \Cref{alg:decsvm} are provided in {Section B} of the Supplementary Material.} The proposed method draws inspiration from algorithms developed in \citet{mateos2010DistributedSparseLinear} and \citet{chang2014MultiAgentDistributedOptimization} for solving $\ell_1$-penalized least squares problems. However, the linear convergence for {\it doubly} nonsmooth objective function is unclear there.

\Cref{alg:decsvm} offers three key advantages. (i) {\it Reduced Storage and Communication Costs:}  The redundant auxiliary variables and multipliers $\{\t^{(\ell k)}, \bu^{(\ell k)}, \bv^{(\ell k)}\}$ are eliminated. Each local node only needs to store and update two $p$-dimensional vectors, $\{\bbeta^{(\ell)}, \bp^{(\ell)}\}$, thereby significantly reducing communication and storage overhead.  (ii) {\it Parallelized Local Updates:} Local updates can be highly parallelized, significantly enhancing computational efficiency.   (iii) {\it Algorithmic Versatility:} The proposed algorithm is easily extensible to other sparsity-promoting penalties, such as the $\ell_0$ penalty, adaptive $\ell_1$ penalty \citep{Zou2006TheAL}, smoothly clipped absolute deviation (SCAD) penalty \citep{fan2001VariableSelectionNonconcave}, and minimax concave penalty (MCP) \citep{zhang2010NearlyUnbiasedVariable}.


We conclude this section with a lemma that establishes the Lipschitz constant \( c_h \) for the kernels under consideration.
\begin{lemm}[Lipschitz Continuity] 
\label{lemma:lipschitz}
The first-order derivatives of the convoluted loss function, instantiated with different types of kernels, \( L_h^\prime(u) \), satisfy  $|L_h^\prime(u_1) - L_h^\prime(u_2)| \leq c_h |u_1 - u_2|,$ 
where the Lipschitz constants \( c_h \) are  $1/(2h)$, $1/(4h)$ and $1/\{(2\pi)^{1/2}h\}$  for  Laplacian, logistic, and Gaussian kernels, respectively.
Consequently, the convoluted hinge loss \( L_h(u) \) satisfies the following quadratic majorization:  
$
	L_h(u_1) \leq L_h(u_2) + L_h^\prime(u_2)(u_1 - u_2) + c_h(u_1 - u_2)^2/2.
$

\end{lemm} 
\begin{algorithm}[!htp]
	\caption{{\small Decentralized penalized Convoluted SVM estimation}}\label{alg:decsvm}
	\begin{algorithmic}[1]
		\REQUIRE Data $\{(\x_i, Y_i)\colon i = 1, \ldots, N\}$, the maximum number of iterations $T$, the regularization parameters  $\lambda_0$ and $\lambda$, the initial estimates $\bhbeta_0^{(\ell)}$ at each node $\ell$.
		\STATE Set $\bbeta_{0}^{(\ell)} = \wh\bbeta_{0}^{(\ell)}$, $\bp_{0}^{(\ell)} = \bm{0}$.
		\FOR{$t = 0, \ldots, T$}
		\STATE Communicate local parameter $\bbeta_{t}^{(\ell)}$ with neighboring nodes;
		\STATE Update $\bbeta_{t}^{(\ell)}$ and $\bp_{t}^{(\ell)}$  with \eqref{update:beta} and \eqref{update:Pbeta}, respectively. 
		\ENDFOR
		\ENSURE {The final estimate 
        $ \bbeta_{t+1}^{(\ell)}$. 
		}
	\end{algorithmic}
\end{algorithm}

\section{Theoretical Properties}\label{section:theory}
Let $\bbeta^\ast_h \defn \argmin_{\bbeta \in \mR^p} \expect\{L_h(Y \x\trans \bbeta)\}$ represent the population parameters under the smoothed loss, and $\bbeta^\ast = (\beta_1^\ast, \ldots, \beta_p^\ast)\trans \defn \argmin_{\bbeta \in \mR^p} \expect\{L(Y \x\trans \bbeta)\}$ be the population parameters under the original hinge loss. 
This study aims to derive a sparse estimate based on the premise that $\bbetaT$ is inherently sparse. Define $\calS = \{j\colon \beta_j^\ast \neq 0, 1 \leq j \leq p\}$ as the support of $\bbetaT$, denoting the indices of significant covariates. Let $s = \abs{\calS}$ signify the number of nonzero elements within $\bbetaT$. We allow both $p = p_N$ and $s = s_N$ to increase with $N$, assuming $s_N \geq 1$ and $p_N \to \infty$ as $N \to \infty$. For simplicity, we use $p$ and $s$ when no confusion is likely.

We now enumerate the assumptions necessary for our theoretical framework. {Recall $\x = (X_1, \ldots, X_p)\trans$ with $X_1\equiv 1$. Let $\x_{-2}$ be a $(p-1)$-dimensional vector with $X_2$ removed from $\x$. Similar notations are used for $\bbeta$. Let $f$ and $g$ be the density functions of $\x$ when $Y = 1$ and $Y=-1$, respectively. Let $f(x\mid \x_{-2})$ be the conditional function of $X_2$ given $\x_{-2}$ and $f_{-2}(\x_{-2})$ be the joint density of $\x_{-2}$. Similar notations are used for $g(\cdot)$. }

\begin{assumptions}
\item \label{condition:network} The peer-to-peer network $\calG$ is connected and has no self-loops.
\item \label{condition:predictor} {The 
predictor $\x_i$ is  sub-exponential, i.e.}, {$\sup_{\abs{\ba}_2 = 1}\expect\{\exp(\abs{\ba\trans\x_i}/m_0)\}\leq 2$} for some $m_0\geq 0$.
\item \label{condition:true_parameter} $\abs{\beta_2^\ast}\geq c$ and $\abs{\bbeta^\ast}_2\leq C$ for some constants $c,C>0$. 
\item \label{condition:rsc}  {Let $\D\defn \expect\big\{ \delta\big(1 - Y\x\trans\bbeta^\ast\big)\x \x\trans\big\}$ be the information matrix of the population hinge loss at the true parameters $\bbeta^\ast$, where $\delta(\cdot)$ is the Dirac delta function. Assume that $\kappa\leq \Lambda_{\min}\{\D\}\leq \Lambda_{\max}\{\D\}\leq \kappa^{-1}$ for some constant $\kappa>0$.  } 
\item \label{condition:kernel} {The kernel function $K\colon \mR\to [0, \infty)$ is  bounded,  
 	Lipschitz continous and symmetric around zero. Also it satisfies $\int_{-\infty}^\infty K(u)\md u = 1$ and $\int_{-\infty}^\infty \exp(tu^2)K(u)\md u \leq C$ for some $t>0$ and $C>0$. 
 	}
\item \label{condition:density} {Assume that $\sup_{x\in\mR} \max\{\abs{f(x\mid \x_{-2})}, \abs{f^\prime(x\mid \x_{-2})}, \abs{x^2f(x\mid \x_{-2})}, \abs{x^2f^\prime(x\mid \x_{-2})}\}\leq C$ for some constant $C>0$. Also assume that $\int_{\mR} \abs{x} f(x\mid \x_{-2})\md x<\infty$. Similar assumptions are made for $g(\cdot)$. }
\item \label{condition:initial} For each $\ell = 1, \ldots, m$,  the initial estimate $\wh\bbeta^{(\ell)}_0$ satisfies $\abs{\wh\bbeta^{(\ell)}_0-\bbetaT}_2 = \O_p(1)$.
\end{assumptions}
\Cref{condition:network} ensures that no subgroup of nodes is isolated, a criterion often referenced in decentralized distributed learning literature. This condition allows all nodes to reach a consensus state upon convergence. \Cref{condition:predictor} pertains to the distribution of the predictors, relaxing the traditional requirement that {$\x$ is bounded} \citep{peng2016ErrorBoundL1norm}. 
\Cref{condition:true_parameter} imposes mild conditions on the boundness of $\bbeta^\ast$. 
\Cref{condition:rsc} introduces eigenvalue conditions on the population Hessian matrix, a common requirement in high-dimensional statistics \citep{wang2019DistributedInferenceLinear,peng2016ErrorBoundL1norm}. \Cref{condition:kernel} is a smoothness { and tail behavior} condition on the kernel function $K(\cdot)$, which is satisfied by commonly used kernel functions. \Cref{condition:density} is a regularity condition on the conditional density $f$ and $g$ and is satisfied by commonly used density functions, e.g., Gaussian distribution and uniform distribution.  In \Cref{alg:decsvm}, the initial estimate is calculated using solely local data at each node. This initial estimate is crafted to meet \Cref{condition:initial}; indeed, even a minimal initial guess, such as setting $\wh\bbeta^{(\ell)}_0 = \bzeros$, is acceptable.

{We first demonstrate the linear convergence of the proposed generalized ADMM in \Cref{alg:decsvm}.
}
{\theo[Linear Convergence]\label{prop:linear_convergence}
	Assume  \Cref{condition:network,condition:initial} and the penalty parameter $\rho_\ell$ satisfies $\rho_\ell\geq c_h \Lambda_{\max}(\sum_{i\in\calI_\ell}\x_i\x_i\trans/n)$ for each $\ell = 1, \ldots, m$, we have {$ (\sum_{\ell = 1}^m\abs{\bbeta^{(\ell)}_{t+ 1} - \wh\bbeta}_2^2)^{1/2} = O_p(\gamma^t)$},   
	where $\gamma\in(0, 1)$. 
}

The convergence factor $\gamma$ is influenced by the network topology $\W$ and the singularity characteristics of the covariance matrices. The selection of the step length, $\rho_\ell$, is straightforward, based on the maximum eigenvalue of the local covariance matrix and the Lipschitz constant, $c_h$, of $L_h^\prime(\cdot)$. 


{The following theorem gives the smoothing bias at the population level. }

\begin{theo}\label{theorem:smoothing_bias}
	Assume  \Cref{condition:network,condition:initial}.  We have $\abs{\bbeta^\ast_h - \bbeta^\ast}_2 = O(h^2)$. 
\end{theo}

The following theorem establishes the statistical convergence rate. Due to the sampling error and optimization error, the smoothing parameter $h$ cannot be made arbitrarily small.

{\theo\label{theorem:convergence_rate}
Assume assumptions in \Cref{prop:linear_convergence} and $\log p/N = o(1)$. {Take the bandwidth $h^2 \asymp (\log p/N)^{1/2}$.} {Choose the tuning parameters such that $4\lambda_0 \abs{\bbetaT}_\infty\lesssim (\log p/N)^{1/2}$.} Then, there exists a 
sufficiently large constant $c_0>0$ such that with the choice $\lambda =c_0 \{\log p/N\}^{1/2}$, the elastic-net 
decentralized penalized CSVM estimate $\bbeta^{(\ell)}_{t+1}$ satisfies
$$
 \abs{\bbeta^{(\ell)}_{t+1} - \bbetaT}_2 = \O_p\{({s\log p}/{N})^{1/2} + \gamma^t\}.
$$
}

Relative to the initial estimate $\wh\bbeta^{(\ell)}_0$, the convergence rate of our refined estimate $\bbeta^{(\ell)}_{t + 1}$ improves to $\max\{(s\log p/N)^{1/2}, \gamma^t\}$. The first term reflects the convergence rate of $\abs{\bhbeta - \bbetaT}_2$, where $\bhbeta$ is the pooled estimate defined in \eqref{optimization:elastic_net} utilizing the full sample. The second term represents the optimization error caused by the generalized ADMM in \Cref{alg:decsvm}.

{ By setting the iteration count $t$ to satisfy $t \geq \log(s\log p/N)/\log \gamma$,  our decentralized estimate $\bbeta^{(\ell)}_{t + 1}$ achieves a minimax convergence rate of $O_p\{(s\log p/N)^{1/2}\}$.}
Notably, this rate is optimal even when all data are centralized at a single node \citep{peng2016ErrorBoundL1norm}. Moreover, our deCSVM demonstrates superior computational efficiency compared to the classical penalized SVM due to the smoothness of its loss function. Additionally, the following corollary shows that $t$ only needs to be of constant order.
\begin{coro}
\label{coro:upper_bound_of_ADMM_iterations}
	Under the assumptions of \Cref{theorem:convergence_rate}, when $p = O\{\exp(N/s)\}$, we further have the number of loops $t\geq \log(s\log p/N)/\log \gamma$ only needs to be constant order.
\end{coro}
The following theorem shows that our deCSVM can exactly recover the true support after sufficient ADMM iterations under some mild conditions. 
\begin{theo}\label{theorem:support_recovery}
	Take $\wh\bbeta^{(\ell)} = \calS_{\lambda}(\bbeta_{t+1}^{(\ell)})$. Assume assumptions in \Cref{prop:linear_convergence}, the irrepresentative condition $\norm{\D_{\calS^c\calS}\D_{\calS\calS}^{-1}}_{\infty}<1 - \alpha$ for some $\alpha\in(0, 1)$, $s^{16/3}\log p/N  = o(1)$ and $\gamma^t\lesssim(\log p/N)^{1/2}$. Denote by $\supp(\wh\bbeta^{(\ell)})$ the support of $\wh\bbeta^{(\ell)}$.
	\begin{enumerate}
		\item[(i)]  We have $\pr\{\supp(\wh\bbeta^{(\ell)})\subseteq \calS\}\to 1$.
		\item[(ii)] Also assume that  $\min_{j\in\calS}\abs{\beta_j^\ast}\geq C\norm{ \D_{\calS\calS}^{-1}}_\infty(\log p/N)^{1/2}$ for some sufficiently large constant $C$. We have $\pr\{\supp(\wh\bbeta^{(\ell)}\} =  \calS)\to 1$. 
	\end{enumerate}
\end{theo}
{
The irrepresentative condition is widely used for establishing support recovery for Lasso-type estimates. The ``beta-min'' condition for exact support recovery matches the one in Lasso with a single machine setting \citep{wainwright2019HighDimensionalStatisticsNonAsymptotic}. 
}

\section{Simulation Studies}\label{section:numerical_studies}
In this section, we conduct extensive simulation studies to examine the finite performance of our proposed deCSVM.

\subsection{Design of the Simulations}
The data generation process is described as follows. A connected network comprising $m$ nodes is generated from an Erd\"{o}s-R\`{e}nyi random graph with a connection probability of $p_c$. At each node, the binary response variable $Y$ is generated such that $\pr(Y = 1) = \pr(Y = -1) = 0.5$. The covariate vector $\x$ is sampled from the multivariate Gaussian distribution $\calN(\bmu_+, \bSig)$ when $Y = 1$ and from $\calN(\bmu_-, \bSig)$ when $Y = -1$. The mean vectors are specified as $\bmu_+ = -\bmu_- = (\mu \bm{1}_s, \bm{0}_{p-s})\trans \in \mR^p$, where $\mu = 0.4$. The covariance matrix $\bSig$ adopts a block diagonal structure, comprising two blocks: $\bSig_{s \times s}$ and $\bSig_{(p-s) \times (p-s)}$, both of which are AR($\rho$) correlation matrices. We vary the correlation parameter $\rho$ in $\{0.3, 0.5, 0.7, 0.9\}$. To accommodate the piratical situations, we randomly flip the sign of the responses with flipping probability $p_{\mathrm{flip}}$. Without further specification, we set parameters $s = 10$, $m = 10$, $p_c = 0.5$, $\sigma^2 = 1$, and $p_{\mathrm{flip}} = 0.01$. Data are generated from the above model independently and identically across the network.


The following lemma provides the exact form of the true separating hyperplane for the population SVM in the absence of sign flips.
{\lemm[True Parameter]\label{lemma:true_parameter}
Define the Mahalanobis distance between $\bmu_+$ and $\bmu_-$ as \( d_{\bSig}(\bmu_+, \bmu_-) = \{ (\bmu_+ - \bmu_-)\trans \bSig^{-1} (\bmu_+ - \bmu_-) \}^{1/2} \). Let \( a^\ast = \gamma^{-1}\{ d_{\bSig}(\bmu_+, \bmu_-)/2 \} \), where \( \gamma(a) = \phi(a)/\Phi(a) \), and \( \phi \) and \( \Phi \) represent the probability density function and cumulative distribution function of the standard Gaussian distribution, respectively. The true parameter vector is given by 
{\( \bbeta^\ast = (\beta_1^\ast, \bbeta_{-}\strans)\trans \), where the intercept term \( \beta_1^\ast = -(\bmu_+ - \bmu_-)\trans \bSig^{-1} (\bmu_+ + \bmu_-)/A \)} 
and the slope vector \( \bbeta_{-}^\ast = 2 \bSig^{-1} (\bmu_+ - \bmu_-)/A \). The normalizing constant \( A \) is defined as \( A = 2 a^\ast d_{\bSig}(\bmu_+, \bmu_-) + d_{\bSig}^2(\bmu_+, \bmu_-) \).
}

In the simulations, our deCSVM is constructed with $\ell_1$ penalty. An Epanechnikov kernel is used except for \Cref{subsection:effect_of_iterations}. 
 At each node, we take the local $\ell_1$-penalized SVM based on local data as initial estimate. We choose the tuning parameter $\lambda$ by minimizing the modified Bayesian information criterion proposed by \citep{zhang2016consistent} and defined as
\beqrs
N^{-1}  \sum_{\ell=1}^{m} \sum_{i \in \calI_\ell} \left(1 - Y_i \x_i\trans \wh{\bbeta}^{(\ell)} \right)_{+} + (\log N)^{1/2} \log p  \sum_{\ell = 1}^m {\supp(\wh\bbeta^{(\ell)})}/{m},
\eeqrs
for any set of candidate estimates $\{\wh\bbeta^{(\ell)}\}_{\ell=1}^m$.  In practice, the use of a gossip protocol allows for efficient broadcasting of scalar values (loss and estimated sparsity) across the network, ensuring that communication costs remain low. 
The bandwidth $h$ is set to $\max\{(\log p/N)^{1/4}, 0.05\}$ according to \Cref{theorem:convergence_rate}.


We compare our deCSVM with the following four competitors: 1) Pooled SVM, which computes the $\ell_1$-penalized SVM estimate using the entire dataset and serves as the benchmark; 2) Local SVM, where each node independently computes its own $\ell_1$-penalized SVM estimate using only the local data available at that node; 3) Average SVM, which aggregates the local SVM estimates by averaging them across nodes, employing the consensus protocol outlined in \citep{yadav2007distributed}; 4) Decentralized subgradient descent (D-subGD), where nodes collaboratively solve \eqref{loss:consensus} with subgradient descent, leveraging information exchanged with neighboring nodes


We evaluate estimation performance using two metrics. The first is the estimation error, defined as \((\sum_{\ell = 1}^m \abs{\wh\bbeta^{(\ell)} - \bbetaT}_2^2 / m )^{1/2}\), which quantifies the average Euclidean distance between the estimated parameter vectors \(\wh\bbeta^{(\ell)}\) and the true parameter vector \(\bbetaT\) across all nodes.   
The second is the mean $F_1$-score. At each node \(\ell\), the \(F_1\)-score is defined as the harmonic mean of precision and recall, where precision is given by 
{
\(|\supp(\wh\bbeta^{(\ell)}) \cap \supp(\bbeta^{*})| / |\supp(\wh\bbeta^{(\ell)})|\) and recall is \(|\supp(\wh\bbeta^{(\ell)}) \cap \supp(\bbeta^{*})| / |\supp(\bbeta^{*})|\). Here, \(\supp(\wh\bbeta^{(\ell)})\) represents the support of \(\wh\bbeta^{(\ell)}\).} 
The \(F_1\)-score ranges from 0 to 1, with higher values indicating superior accuracy in support recovery. The communication budget is set to $100$ rounds for all decentralized methods. All reported values are averaged over $100$ independent replications. 

\subsection{Effect of Iterations}\label{subsection:effect_of_iterations}
\begin{figure}[htp]
	\centerline{
		\begin{tabular}{cc}
\psfig{figure=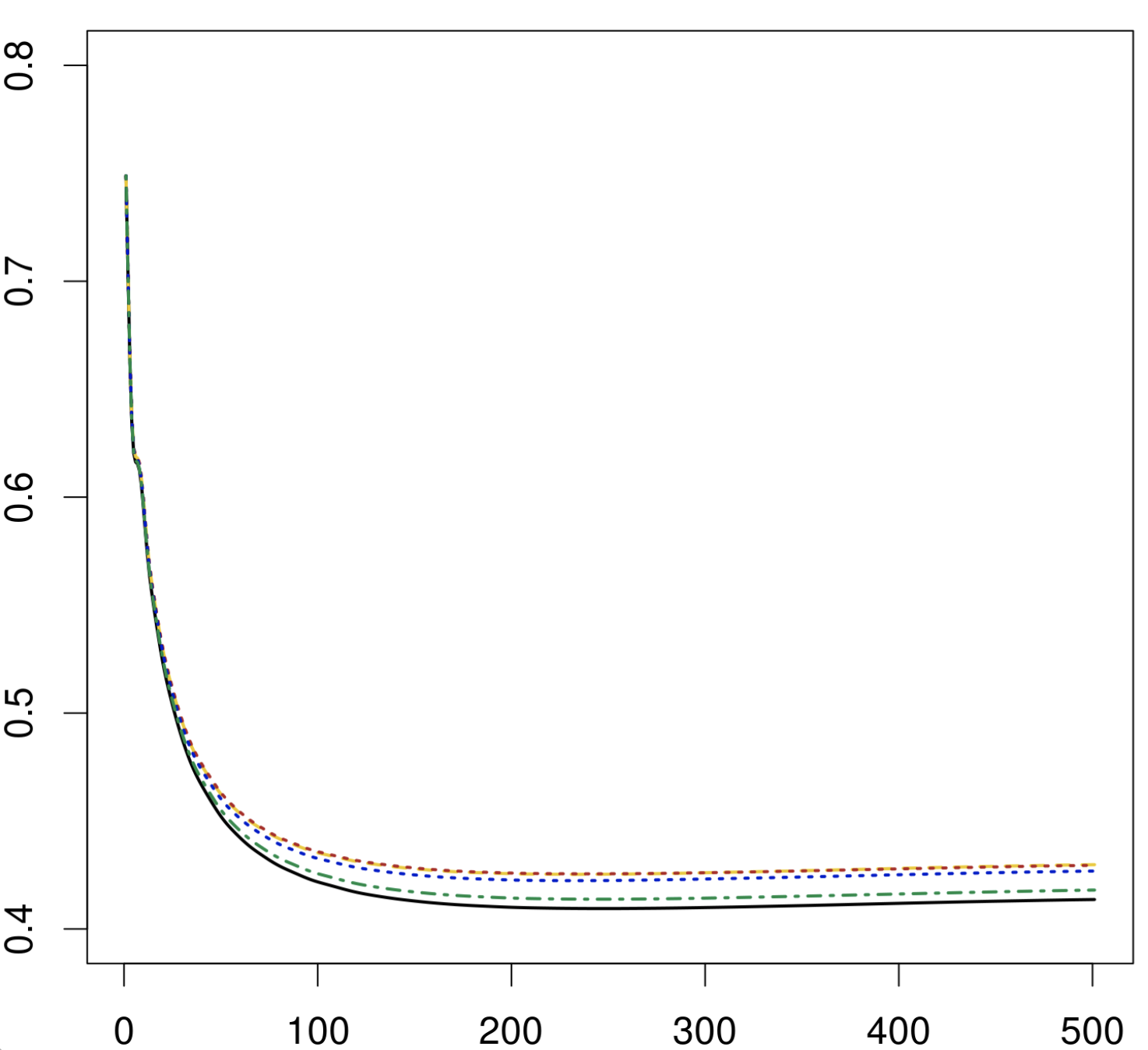,width=3in,angle=0}
      & \psfig{figure=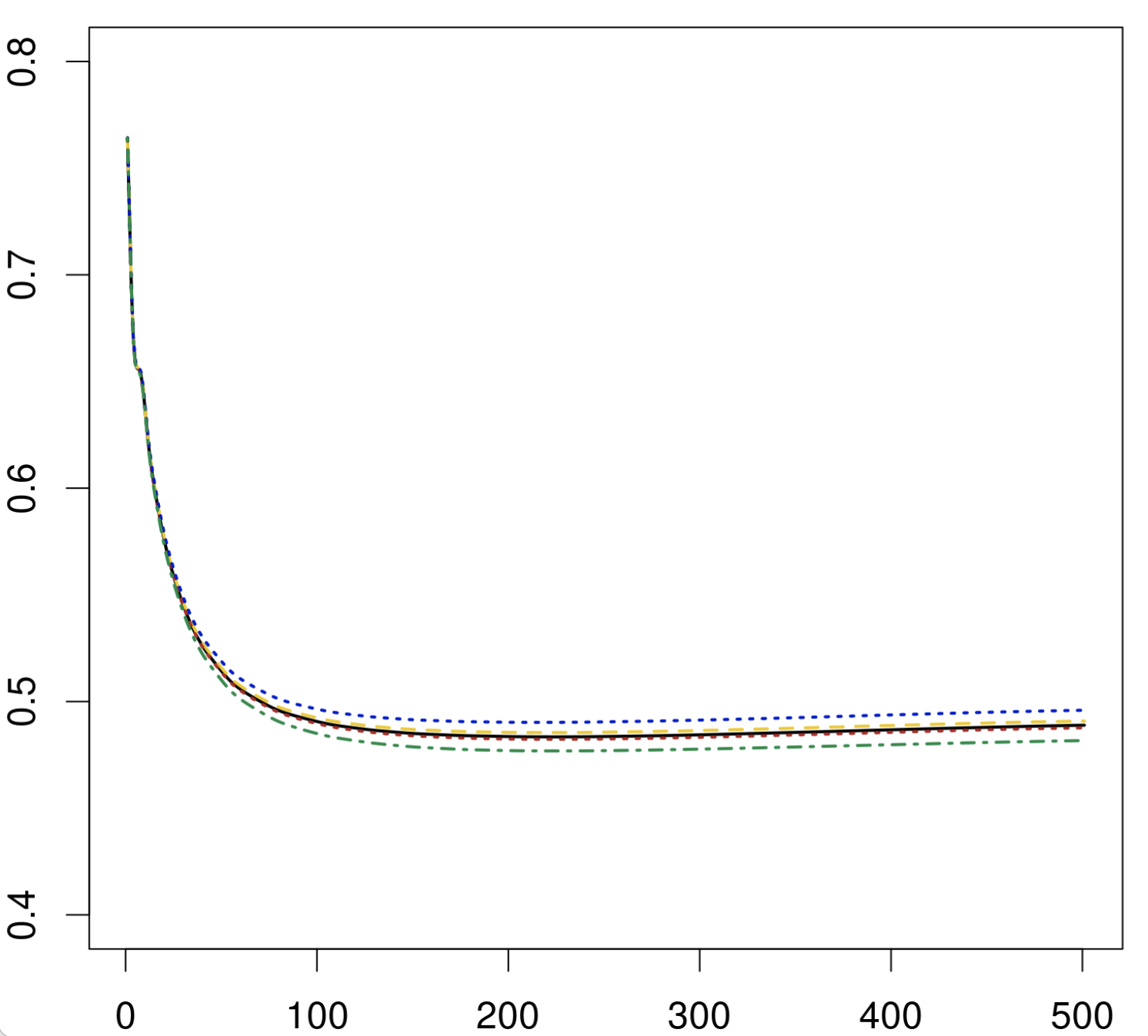,width=3in,angle=0} \\

 		(a): $p=50$ and $n = 100$  & (b): $p =100$ and $n = 200$ \\
		\end{tabular}
	}
	\captionsetup{font = footnotesize}
	\caption{ The horizontal axis stands for the number of  iterations, and the vertical axis represents the estimation error of the deCSVM estimate under different kernel types:  uniform (\uniformline), Laplacian (\laplacianline), logistic (\logisticline), Gaussian (\gaussianline), and Epanechnikov (\epanechnikovline).} 
	\label{fig:iterations}
\end{figure}

We evaluate the impact of the number of iterations on the performance of the proposed deCSVM method using various kernel functions: uniform, Laplacian, logistic, Gaussian, and Epanechnikov. Two experimental settings are considered: (a) \( p = 50 \), \( n = 100 \) and (b) \( p = 100 \), \( n = 200 \). The estimation errors as a function of the iteration count are depicted in \Cref{fig:iterations}. The results demonstrate a clear linear decline in estimation errors as the iterations progress, stabilizing after approximately $200$ iterations. This behavior aligns with the theoretical findings presented in \Cref{prop:linear_convergence}. Furthermore, the stabilized error values are similar across different kernels, underscoring the robustness of deCSVM to the choice of kernel function.


\graphicspath{{figs/}}

\subsection{Effect of Local Sample Size}
In this subsection, we investigate the impact of local sample size on the performance of the proposed deCSVM method. Specifically, we vary the local sample size \(n\) and the dimensionality \(p\) across the configurations \(\{(100, 100), (200, 100), (200, 200)\}\). The estimation errors and \(F_1\)-scores for the competing methods are presented in \Cref{tab:local_samp_err,tab:local_samp_f1}, respectively. The results demonstrate that deCSVM achieves estimation errors and \(F_1\)-scores comparable to the benchmark pooled estimator while outperforming other decentralized methods. Notably, deCSVM is the only decentralized approach that consistently attains high \(F_1\)-scores, underscoring its effectiveness in generating sparse classification rules.


\begin{table}[ht]
	\footnotesize
	\captionsetup{font = footnotesize}
	\caption{Estimation errors of the Pooled, Local, Avg., D-subGD, and deCSVM estimates for varying local sample sizes and the dimensions $(n, p)$ in $\{(100, 100), (200, 100), (200, 200)\}$, under different values of $\rho$.   }
 \label{tab:local_samp_err}
	\centering {\footnotesize
\begin{tabular}{c|c|cccc}
\hline
$(n, p)$ & Pooled & Local & Avg. & D-subGD & deCSVM \\
\hline
& \multicolumn{5}{c}{$\rho = 0.3$} \\
\cline{2-6}
(100, 100) & 0.4730 & 0.9254 & 0.6863 & 0.5938 & 0.5523 \\ 
(200, 100) & 0.3874 & 0.7512 & 0.5797 & 0.4558 & 0.4371 \\ 
(200, 200) & 0.4069 & 0.7759 & 0.6295 & 0.5204 & 0.4817 \\ 
\hline
& \multicolumn{5}{c}{$\rho = 0.5$} \\
\cline{2-6}
(100, 100) & 0.3681 & 0.8215 & 0.5237 & 0.6518 & 0.4700 \\ 
(200, 100) & 0.2931 & 0.6486 & 0.4262 & 0.5068 & 0.3239 \\ 
(200, 200) & 0.2979 & 0.6632 & 0.4639 & 0.5812 & 0.3610 \\ 
\hline
& \multicolumn{5}{c}{$\rho = 0.7$} \\
\cline{2-6}
(100, 100) & 0.3582 & 0.8007 & 0.4490 & 0.7682 & 0.4185 \\ 
(200, 100) & 0.2631 & 0.6071 & 0.3511 & 0.5777 & 0.3102 \\ 
(200, 200) & 0.2639 & 0.6142 & 0.3725 & 0.6648 & 0.3307 \\ 
\hline
& \multicolumn{5}{c}{$\rho = 0.9$} \\
\cline{2-6}
(100, 100) & 0.3605 & 0.9296 & 0.4856 & 1.0392 & 0.4426 \\ 
(200, 100) & 0.3076 & 0.7145 & 0.4066 & 0.7600 & 0.3812 \\ 
(200, 200) & 0.2978 & 0.6885 & 0.4189 & 0.8438 & 0.4010 \\ 
\hline
\end{tabular}}
\end{table}


\begin{table}[ht]
	\footnotesize
	\captionsetup{font = footnotesize}
	\caption{$F_1$-scores of the Pooled, Local, Avg., D-subGD, and deCSVM estimates for varying local sample sizes and the dimensions $(n, p)$ in $\{(100, 100), (200, 100), (200, 200)\}$, under different values of $\rho$.  }
 \label{tab:local_samp_f1}
	\centering {\footnotesize
\begin{tabular}{c|c|cccc}
\hline
$(n, p)$ & Pooled & Local & Avg. & D-subGD & deCSVM \\
\hline
& \multicolumn{5}{c}{$\rho = 0.3$} \\
\cline{2-6}
(100, 100) & 0.9469 & 0.4974 & 0.2198 & 0.1818 & 0.8184 \\ 
(200, 100) & 0.9485 & 0.5863 & 0.2184 & 0.1818 & 0.7913 \\ 
(200, 200) & 0.9451 & 0.5539 & 0.1600 & 0.0952 & 0.8176 \\ 
\hline
& \multicolumn{5}{c}{$\rho = 0.5$} \\
\cline{2-6}
(100, 100) & 0.9326 & 0.4549 & 0.2385 & 0.1818 & 0.8613 \\ 
(200, 100) & 0.9472 & 0.5558 & 0.2382 & 0.1818 & 0.8053 \\ 
(200, 200) & 0.9485 & 0.5288 & 0.1760 & 0.0952 & 0.8263 \\ 
\hline
& \multicolumn{5}{c}{$\rho = 0.7$} \\
\cline{2-6}
(100, 100) & 0.8087 & 0.3847 & 0.2602 & 0.1818 & 0.8715 \\ 
(200, 100) & 0.8941 & 0.4943 & 0.2787 & 0.1818 & 0.8442 \\ 
(200, 200) & 0.8977 & 0.4633 & 0.2049 & 0.0952 & 0.8556 \\ 
\hline
& \multicolumn{5}{c}{$\rho = 0.9$} \\
\cline{2-6}
(100, 100) & 0.5744 & 0.2782 & 0.2823 & 0.1818 & 0.8736 \\ 
(200, 100) & 0.6723 & 0.3777 & 0.3053 & 0.1818 & 0.8682 \\ 
(200, 200) & 0.6566 & 0.3493 & 0.2478 & 0.0952 & 0.8802 \\ 
\hline
\end{tabular}}
\end{table}


\subsection{Effect of Topology}
We investigate the impact of network topology on the performance of our deCSVM method, focusing on two key factors: the number of nodes and network sparsity.

To evaluate the effect of the number of nodes, we fix the total sample size to \(N = 4000\) and vary the number of nodes \(m \in \{5, 10, 20\}\). To eliminate network randomness, we consider a fully connected decentralized network, i.e., \(p_c = 1\). The results, summarized in \Cref{table:number_of_nodes}, demonstrate that deCSVM exhibits robustness to the number of nodes, achieving consistent performance across all configurations. Notably, deCSVM remains comparable to the pooled estimator and consistently outperforms other decentralized methods.


Next, we assess the effect of network sparsity by varying the network connection probability \(p_c \in \{0.3, 0.5, 0.8\}\). For this analysis, we fix the number of nodes to \(m = 10\), the local sample size to \(n = 200\), and the dimensionality to \(p = 100\). As shown in \Cref{table:probability_of_connection}, deCSVM is relatively insensitive to network sparsity. This observation aligns with our theoretical result in \Cref{prop:linear_convergence}, which states that network sparsity impacts only the convergence factor \(\gamma\); with sufficient ADMM iterations, deCSVM achieves the optimal statistical convergence rate. In contrast, the D-subGD method exhibits sensitivity to network sparsity, particularly under settings with \(\rho = 0.9\). This maybe due to its slow convergence rate.

\begin{table}[ht]
\captionsetup{font = footnotesize}
\caption{Estimation errors and $F_1$-scores of the Pooled, Local, Avg., D-subGD, and deCSVM estimates for varying the number of nodes $m$ in $\{5, 10, 20\}$ under different $\rho$ values.}
\label{table:number_of_nodes}
\centering
\resizebox{\columnwidth}{!}{\footnotesize
\begin{tabular}{c|cc|cc|cc|cc|cc}
\hline
$m$ & \multicolumn{2}{c|}{Pooled} & \multicolumn{2}{c|}{Local} & \multicolumn{2}{c|}{Avg.} & \multicolumn{2}{c|}{D-subGD} & \multicolumn{2}{c}{deCSVM} \\
\hline
& Est. error & $F_1$-score & Est. error & $F_1$-score & Est. error & $F_1$-score & Est. error & $F_1$-score & Est. error & $F_1$-score \\
\cline{2-11}
& \multicolumn{10}{c}{$\rho = 0.3$} \\
\cline{2-11}
5  & 0.3103 & 0.9442 & 0.4299 & 0.5677 & 0.3388 & 0.2576 & 0.2972 & 0.1818 & 0.2838 & 0.8602 \\ 
10 & 0.3121 & 0.9472 & 0.5561 & 0.5852 & 0.4111 & 0.2069 & 0.3240 & 0.1818 & 0.2984 & 0.7666 \\ 
20 & 0.3197 & 0.9489 & 0.7130 & 0.5761 & 0.4969 & 0.1856 & 0.4161 & 0.1818 & 0.3741 & 0.8091 \\ 
\hline
& \multicolumn{10}{c}{$\rho = 0.5$} \\
\cline{2-11}
5  & 0.2209 & 0.9498 & 0.3787 & 0.6314 & 0.2511 & 0.3129 & 0.3745 & 0.1818 & 0.2021 & 0.8571 \\ 
10 & 0.2203 & 0.9498 & 0.4865 & 0.6100 & 0.2929 & 0.2274 & 0.3938 & 0.1818 & 0.2054 & 0.7706 \\ 
20 & 0.2264 & 0.9515 & 0.6258 & 0.5683 & 0.3581 & 0.1920 & 0.4191 & 0.1818 & 0.2554 & 0.8080 \\ 
\hline
& \multicolumn{10}{c}{$\rho = 0.7$} \\
\cline{2-11}
5  & 0.1971 & 0.9403 & 0.3609 & 0.6338 & 0.2041 & 0.3764 & 0.4216 & 0.1818 & 0.1787 & 0.8653 \\ 
10 & 0.1954 & 0.9373 & 0.4689 & 0.5822 & 0.2431 & 0.2670 & 0.4307 & 0.1818 & 0.1926 & 0.7731 \\ 
20 & 0.2058 & 0.9310 & 0.5958 & 0.5177 & 0.2967 & 0.2067 & 0.4691 & 0.1818 & 0.2319 & 0.8024 \\ 
\hline
& \multicolumn{10}{c}{$\rho = 0.9$} \\
\cline{2-11}
5  & 0.2255 & 0.7372 & 0.4032 & 0.5145 & 0.2466 & 0.4523 & 0.5280 & 0.1818 & 0.2330 & 0.8784 \\ 
10 & 0.2346 & 0.7232 & 0.5219 & 0.4668 & 0.2962 & 0.3342 & 0.5739 & 0.1818 & 0.2868 & 0.8666 \\ 
20 & 0.2239 & 0.7356 & 0.6806 & 0.3936 & 0.3500 & 0.2310 & 0.5981 & 0.1818 & 0.3277 & 0.7551 \\ 
\hline
\end{tabular}}
\end{table}

\begin{table}[ht]
\captionsetup{font = footnotesize}
\caption{Estimation errors and $F_1$-scores of the Pooled, Local, Avg., D-subGD, and deCSVM estimates for varying $p_c$ in $\{0.3, 0.5, 0.8\}$ under different $\rho$ values.}
\label{table:probability_of_connection}
\centering
    \resizebox{\columnwidth}{!}{\footnotesize
\begin{tabular}{c|cc|cc|cc|cc|cc}
\hline
$p_c$ & \multicolumn{2}{c|}{Pooled} & \multicolumn{2}{c|}{Local} & \multicolumn{2}{c|}{Avg.} & \multicolumn{2}{c|}{D-subGD} & \multicolumn{2}{c}{deCSVM} \\
\hline
& Est. error & $F_1$-score & Est. error & $F_1$-score & Est. error & $F_1$-score & Est. error & $F_1$-score & Est. error & $F_1$-score \\
\cline{2-11}
& \multicolumn{10}{c}{$\rho = 0.3$} \\
\cline{2-11}
0.3 & 0.3861 & 0.9468 & 0.7130 & 0.5787 & 0.5125 & 0.2115 & 0.4325 & 0.1818 & 0.3558 & 0.7923 \\ 
0.5 & 0.3849 & 0.9481 & 0.7156 & 0.5787 & 0.5129 & 0.2110 & 0.4298 & 0.1818 & 0.3604 & 0.7879 \\ 
0.8 & 0.3913 & 0.9498 & 0.7100 & 0.5783 & 0.5155 & 0.2115 & 0.4194 & 0.1818 & 0.3730 & 0.7891 \\ 
\hline
& \multicolumn{10}{c}{$\rho = 0.5$} \\
\cline{2-11}
0.3 & 0.2807 & 0.9477 & 0.6292 & 0.5624 & 0.3745 & 0.2304 & 0.5050 & 0.1818 & 0.2533 & 0.7770 \\ 
0.5 & 0.2911 & 0.9482 & 0.6348 & 0.5802 & 0.3843 & 0.2396 & 0.5010 & 0.1818 & 0.2672 & 0.7981 \\ 
0.8 & 0.2832 & 0.9490 & 0.6250 & 0.5644 & 0.3757 & 0.2332 & 0.4818 & 0.1818 & 0.2709 & 0.7875 \\ 
\hline
& \multicolumn{10}{c}{$\rho = 0.7$} \\
\cline{2-11}
0.3 & 0.2693 & 0.8864 & 0.5876 & 0.5155 & 0.3181 & 0.2660 & 0.5726 & 0.1818 & 0.2732 & 0.8265 \\ 
0.5 & 0.2657 & 0.8891 & 0.5983 & 0.5188 & 0.3199 & 0.2683 & 0.5659 & 0.1818 & 0.2722 & 0.8343 \\ 
0.8 & 0.2730 & 0.8903 & 0.5961 & 0.5133 & 0.3225 & 0.2704 & 0.5689 & 0.1818 & 0.2731 & 0.8233 \\ 
\hline
& \multicolumn{10}{c}{$\rho = 0.9$} \\
\cline{2-11}
0.3 & 0.2893 & 0.6737 & 0.6818 & 0.3937 & 0.3823 & 0.3115 & 0.7693 & 0.1818 & 0.3623 & 0.8665 \\ 
0.5 & 0.2887 & 0.6586 & 0.6777 & 0.3983 & 0.3830 & 0.3244 & 0.7227 & 0.1818 & 0.3623 & 0.8683 \\ 
0.8 & 0.2961 & 0.6541 & 0.6749 & 0.3960 & 0.3755 & 0.3218 & 0.6967 & 0.1818 & 0.3552 & 0.8654 \\ 
\hline
\end{tabular}}
\end{table}

\subsection{Effect of Sign Flips}\label{subsection:signflip}
In this subsection, we evaluate the performance of our deCSVM under different flipping probability $p_{\text{flip}} \in \{0.01, 0.05, 0.1\}$. These levels represent varying degrees of misclassification, which are commonly found in real-world datasets due to errors in data labeling, collection, or processing. 


 \Cref{table:sign_flip} shows that as flipping probability increases, all methods experience higher estimation errors and lower $F_1$-scores. However, our method exhibits greater resilience, with smallest estimation error and less reduction in $F_1$-score among decentralized methods. This demonstrates that our method maintains reliable estimation and support recovery even under significant label noise, highlighting its robustness in real-world scenarios with imperfect data.



\begin{table}[ht]
\captionsetup{font=footnotesize}
\caption{Estimation error and $F_1$-scores of different methods for  varying $p_{\mathrm{flip}}$ in $\{0.01, 0.05, 0.1\}$ under different $\rho$ values.}
\label{table:sign_flip}
\centering
    \resizebox{\columnwidth}{!}{\footnotesize
\begin{tabular}{c|cc|cc|cc|cc|cc}
\hline
$p_{\mathrm{flip}}$ & \multicolumn{2}{c|}{Pooled} & \multicolumn{2}{c|}{Local} & \multicolumn{2}{c|}{Avg.} & \multicolumn{2}{c|}{D-subGD} & \multicolumn{2}{c}{deCSVM} \\
\hline
& Est. error & $F_1$-score & Est. error & $F_1$-score & Est. error & $F_1$-score & Est. error & $F_1$-score & Est. error & $F_1$-score \\
\cline{2-11}
& \multicolumn{10}{c}{$\rho = 0.3$} \\
\cline{2-11}
0.3 & 0.3861 & 0.9468 & 0.7205 & 0.5732 & 0.5096 & 0.2089 & 0.4204 & 0.1818 & 0.3613 & 0.8001 \\ 
0.5 & 0.4570 & 0.9498 & 0.7479 & 0.5921 & 0.5798 & 0.2219 & 0.4563 & 0.1818 & 0.4268 & 0.7932 \\ 
0.7 & 0.5326 & 0.9497 & 0.8022 & 0.5854 & 0.6526 & 0.2346 & 0.5303 & 0.1818 & 0.5074 & 0.7951 \\ 
\hline
& \multicolumn{10}{c}{$\rho = 0.5$} \\
\cline{2-11}
0.3 & 0.2807 & 0.9477 & 0.6348 & 0.5618 & 0.3738 & 0.2281 & 0.4944 & 0.1818 & 0.2565 & 0.7910 \\ 
0.5 & 0.3334 & 0.9483 & 0.6460 & 0.5670 & 0.4259 & 0.2416 & 0.5106 & 0.1818 & 0.3195 & 0.8023 \\ 
0.7 & 0.3840 & 0.9353 & 0.6791 & 0.5374 & 0.4758 & 0.2460 & 0.5263 & 0.1818 & 0.3878 & 0.8134 \\ 
\hline
& \multicolumn{10}{c}{$\rho = 0.7$} \\
\cline{2-11}
0.3 & 0.2693 & 0.8864 & 0.5971 & 0.5121 & 0.3212 & 0.2658 & 0.5770 & 0.1818 & 0.2765 & 0.8379 \\ 
0.5 & 0.2870 & 0.8789 & 0.6173 & 0.4928 & 0.3532 & 0.2691 & 0.5761 & 0.1818 & 0.3052 & 0.8344 \\ 
0.7 & 0.3387 & 0.8267 & 0.6526 & 0.4612 & 0.3998 & 0.2732 & 0.5900 & 0.1818 & 0.3629 & 0.8706 \\ 
\hline
& \multicolumn{10}{c}{$\rho = 0.9$} \\
\cline{2-11}
0.3 & 0.2893 & 0.6737 & 0.6702 & 0.3944 & 0.3813 & 0.3206 & 0.7352 & 0.1818 & 0.3602 & 0.8805 \\ 
0.5 & 0.2969 & 0.6426 & 0.7000 & 0.3796 & 0.4013 & 0.3233 & 0.7255 & 0.1818 & 0.3802 & 0.8728 \\ 
0.7 & 0.3441 & 0.6146 & 0.7237 & 0.3506 & 0.4314 & 0.3067 & 0.7764 & 0.1818 & 0.4096 & 0.8740 \\ 
\hline
\end{tabular}}
\end{table}

\section{An Application}\label{section:application}
\begin{figure}[htp]
\centering
\includegraphics[scale = 0.5]{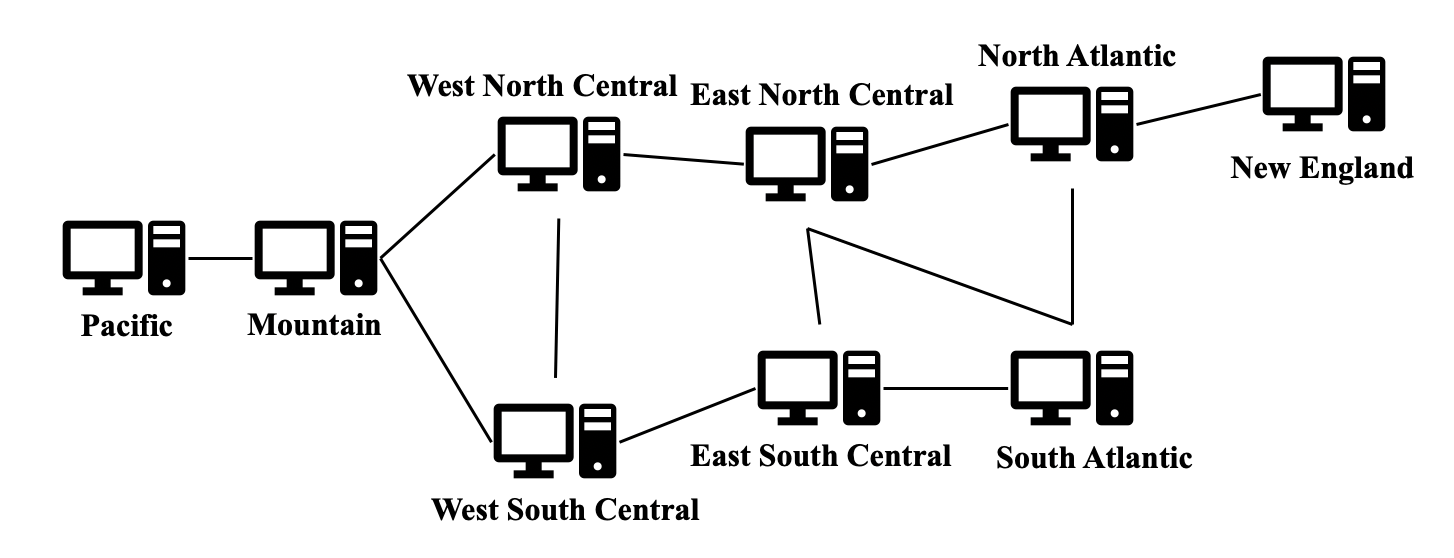}
	\captionsetup{font = footnotesize}
	\caption{Decentralized network of the communities and crime dataset.} 
	\label{fig:real}
\end{figure}

We apply our proposed deCSVM method to classify communities into high- and low-crime categories. The dataset used is the Communities and Crime dataset from the UCI Machine Learning Repository, which is publicly accessible at \url{https://archive.ics.uci.edu/dataset/183/communities+and+crime}. This dataset contains 147 variables describing 2,215 communities across 49 U.S. states. A community is classified as high crime if its crime rate exceeds the median value of \(0.15\); otherwise, it is labeled as low crime. The communities are grouped into nine nodes based on Census Bureau-designated divisions, as depicted in \Cref{fig:real}. Network connections are determined by spatial proximity, reflecting real-world constraints such as geographic adjacency, communication limitations, and privacy requirements. This decentralized structure promotes efficient, region-specific data sharing while maintaining local autonomy.

After removing variables with missing values and normalizing the data, the final dataset comprises 99 variables and 1,993 communities. To simulate noisy labels, we introduce sign flips to the response variable with varying flipping probabilities \(p_{\text{flip}} \in \{0, 0.01, 0.05\}\). The dataset is then randomly split into training and testing sets using an 8:2 ratio. 

We compare the performance of deCSVM with the most competitive decentralized method, D-subGD. Classification accuracy on the testing set and the mean support size are evaluated over 100 independent random splits, as summarized in \Cref{table:flip_accuracy_dimension}. The results indicate that deCSVM achieves higher classification accuracy while yielding a significantly smaller support set. The sparsity of the resulting classification rule enhances its interpretability, providing clearer insights into the decision-making process.

{

\begin{table}[ht]
\captionsetup{font=footnotesize}
\caption{Classification accuracy on the testing set and the mean support size for the D-subGD and deCSVM methods are evaluated under varying flipping probabilities \(p_{\text{flip}}\), based on 100 independent random splits.}
\label{table:flip_accuracy_dimension}
\centering
\footnotesize  
\begin{tabular}{c|cc|cc}
\hline
$p_{\rm flip}$ & \multicolumn{2}{c|}{D-subGD} & \multicolumn{2}{c}{deCSVM} \\
\hline
& Accuracy & Mean support size & Accuracy &  Mean support size\\
\cline{2-5}
0   & 0.8077 & 99 & 0.8287 & 30.4656 \\ 
0.01 & 0.8113 & 99 & 0.8226 & 29.8133 \\ 
0.05 & 0.8105 & 99 & 0.8221 & 28.4878 \\ 
\hline
\end{tabular}
\end{table}
}

\section{Conclusion}\label{section:conclusion}
We propose decentralized convoluted support vector machines for efficiently learning sparse classification rules over networks. By applying convolution to the non-smooth hinge loss, we transform it into a smooth and convex loss function. This modification facilitates the development of a linearly convergent algorithm with straightforward implementation, even when communication is constrained to local neighborhoods. Our work can be extended in three key directions. First, while the current approach focuses on the \(\ell_1\) and elastic-net penalties, its inherent flexibility allows for the incorporation of general nonconvex penalties via a straightforward linear approximation \citep{zou2008OnestepSparseEstimates}. Second, in many practical applications, a substantial amount of unlabeled auxiliary data is available. Leveraging these data has the potential to significantly improve classification accuracy. Third, an important direction for future research is the development of methods for conducting statistical inference on the learned classification rules.


\bigskip
\begin{center}
{\large\bf Supplementary Material}
\end{center}
{
\begin{description}

\item[R-package for deCSVM:] R-package deCSVM contains code to replicate all numerical results. (GNU zipped tar file)

\item[Supplementary Material:] Supplementary Material for ``Efficient Distributed Learning over Decentralized Networks with Convoluted Support Vector Machine'' contains all technical details and additional simulations. (.pdf file)
\end{description}}



\bibliographystyle{Chicago}

\bibliography{ref}

\begin{thebibliography}{}

\bibitem[\protect\citeauthoryear{{1000 Genomes Project Consortium}, Auton,
  Brooks, Durbin, Garrison, Kang, Korbel, Marchini, McCarthy, McVean, and
  Abecasis}{{1000 Genomes Project Consortium}
  et~al.}{2015}]{1000genomesprojectconsortium2015GlobalReferenceHuman}
{1000 Genomes Project Consortium}, A.~Auton, L.~D. Brooks, R.~M. Durbin, E.~P.
  Garrison, H.~M. Kang, J.~O. Korbel, J.~L. Marchini, S.~McCarthy, G.~A.
  McVean, and G.~R. Abecasis (2015, October).
\newblock A global reference for human genetic variation.
\newblock {\em Nature\/}~{\em 526\/}(7571), 68--74.

\bibitem[\protect\citeauthoryear{Boser, Guyon, and Vapnik}{Boser
  et~al.}{1992}]{boser1992TrainingAlgorithmOptimal}
Boser, B.~E., I.~M. Guyon, and V.~N. Vapnik (1992).
\newblock A training algorithm for optimal margin classifiers.
\newblock In {\em Proceedings of the Fifth Annual Workshop on {{Computational}}
  Learning Theory - {{COLT}} '92}, {Pittsburgh, Pennsylvania, United States},
  pp.\  144--152. {ACM Press}.

\bibitem[\protect\citeauthoryear{Boyd}{Boyd}{2010}]{boyd2010DistributedOptimizationStatistical}
Boyd, S. (2010).
\newblock Distributed optimization and statistical learning via the alternating
  direction method of multipliers.
\newblock {\em Foundations and Trends{\textregistered} in Machine
  Learning\/}~{\em 3\/}(1), 1--122.

\bibitem[\protect\citeauthoryear{Cao, Yu, Ren, and Chen}{Cao
  et~al.}{2013}]{cao2013OverviewRecentProgress}
Cao, Y., W.~Yu, W.~Ren, and G.~Chen (2013).
\newblock An {{Overview}} of {{Recent Progress}} in the {{Study}} of
  {{Distributed Multi-Agent Coordination}}.
\newblock {\em IEEE Transactions on Industrial Informatics\/}~{\em 9\/}(1),
  427--438.

\bibitem[\protect\citeauthoryear{Chang, Hong, and Wang}{Chang
  et~al.}{2014}]{chang2014MultiAgentDistributedOptimization}
Chang, T.-H., M.~Hong, and X.~Wang (2014, February).
\newblock Multi-{{Agent Distributed Optimization}} via {{Inexact Consensus
  ADMM}}.
\newblock {\em IEEE Transactions on Signal Processing\/}~{\em 63}.

\bibitem[\protect\citeauthoryear{Chen and Chen}{Chen and
  Chen}{2024}]{chen2022CommunicationEfficientDistributedSupportVectorMachine}
Chen, B. and C.~Chen (2024).
\newblock Convoluted support matrix machine in high dimensions.
\newblock {\em Statistica Sinica\/}.

\bibitem[\protect\citeauthoryear{Chen and Zhu}{Chen and
  Zhu}{2023}]{chen2023DistributedDecodingHeterogeneous}
Chen, C. and L.~Zhu (2023).
\newblock Distributed {{Decoding From Heterogeneous}} 1-{{Bit Compressive
  Measurements}}.
\newblock {\em Journal of Computational and Graphical Statistics\/}~{\em
  32\/}(3), 884--894.

\bibitem[\protect\citeauthoryear{Chen, Liu, and Zhang}{Chen
  et~al.}{2019}]{chen2019QuantileRegressionMemory}
Chen, X., W.~Liu, and Y.~Zhang (2019, December).
\newblock Quantile regression under memory constraint.
\newblock {\em The Annals of Statistics\/}~{\em 47\/}(6), 3244--3273.

\bibitem[\protect\citeauthoryear{Fan and Li}{Fan and
  Li}{2001}]{fan2001VariableSelectionNonconcave}
Fan, J. and R.~Li (2001, December).
\newblock Variable selection via nonconcave penalized likelihood and its oracle
  properties.
\newblock {\em Journal of the American Statistical Association\/}~{\em
  96\/}(456), 1348--1360.

\bibitem[\protect\citeauthoryear{Fernandes, Guerre, and Horta}{Fernandes
  et~al.}{2021}]{fernandes2021SmoothingQuantileRegressions}
Fernandes, M., E.~Guerre, and E.~Horta (2021, January).
\newblock Smoothing {{Quantile Regressions}}.
\newblock {\em Journal of Business \& Economic Statistics\/}~{\em 39\/}(1),
  338--357.

\bibitem[\protect\citeauthoryear{Friedman, Hastie, and Tibshirani}{Friedman
  et~al.}{2010}]{friedman2010RegularizationPathsGeneralized}
Friedman, J., T.~Hastie, and R.~Tibshirani (2010).
\newblock Regularization paths for generalized linear models via coordinate
  descent.
\newblock {\em Journal of Statistical Software\/}~{\em 33\/}(1), 1.

\bibitem[\protect\citeauthoryear{Friedrichs}{Friedrichs}{1944}]{friedrichs1944IdentityWeakStrong}
Friedrichs, K.~O. (1944).
\newblock The {{Identity}} of {{Weak}} and {{Strong Extensions}} of
  {{Differential Operators}}.
\newblock {\em Transactions of the American Mathematical Society\/}~{\em
  55\/}(1), 132--151.

\bibitem[\protect\citeauthoryear{Galvao and Kato}{Galvao and
  Kato}{2016}]{galvao2016SmoothedQuantileRegression}
Galvao, A.~F. and K.~Kato (2016, July).
\newblock Smoothed quantile regression for panel data.
\newblock {\em Journal of Econometrics\/}~{\em 193\/}(1), 92--112.

\bibitem[\protect\citeauthoryear{He, Pan, Tan, and Zhou}{He
  et~al.}{2023}]{he2021SmoothedQuantileRegression}
He, X., X.~Pan, K.~M. Tan, and W.-X. Zhou (2023).
\newblock Smoothed quantile regression with large-scale inference.
\newblock {\em Journal of Econometrics\/}~{\em 232\/}(2), 367--388.

\bibitem[\protect\citeauthoryear{Hector and Song}{Hector and
  Song}{2020}]{hector2020DoublyDistributedSupervised}
Hector, E.~C. and P.~X.-K. Song (2020).
\newblock Doubly distributed supervised learning and inference with
  high-dimensional correlated outcomes.
\newblock {\em The Journal of Machine Learning Research\/}~{\em 21\/}(1),
  173:6983--173:7017.

\bibitem[\protect\citeauthoryear{Hector and Song}{Hector and
  Song}{2021}]{hector2021DistributedIntegratedMethod}
Hector, E.~C. and P.~X.-K. Song (2021, April).
\newblock A {{Distributed}} and {{Integrated Method}} of {{Moments}} for
  {{High-Dimensional Correlated Data Analysis}}.
\newblock {\em Journal of the American Statistical Association\/}~{\em
  116\/}(534), 805--818.

\bibitem[\protect\citeauthoryear{Horowitz}{Horowitz}{1998}]{horowitz1998BootstrapMethodsMedian}
Horowitz, J.~L. (1998).
\newblock Bootstrap {{Methods}} for {{Median Regression Models}}.
\newblock {\em Econometrica\/}~{\em 66\/}(6), 1327.

\bibitem[\protect\citeauthoryear{Li, Artemiou, and Li}{Li
  et~al.}{2011}]{li2011PrincipalSupportVector}
Li, B., A.~Artemiou, and L.~Li (2011, December).
\newblock Principal support vector machines for linear and nonlinear sufficient
  dimension reduction.
\newblock {\em Annals of Statistics\/}~{\em 39\/}(6), 3182--3210.

\bibitem[\protect\citeauthoryear{Li, Shi, and Yan}{Li
  et~al.}{2019}]{li2019DecentralizedProximalGradientMethod}
Li, Z., W.~Shi, and M.~Yan (2019, September).
\newblock A {{Decentralized Proximal-Gradient Method With Network Independent
  Step-Sizes}} and {{Separated Convergence Rates}}.
\newblock {\em IEEE Transactions on Signal Processing\/}~{\em 67\/}(17),
  4494--4506.

\bibitem[\protect\citeauthoryear{Lian and Fan}{Lian and
  Fan}{2018}]{lian2018DivideandConquerDebiasedNorm}
Lian, H. and Z.~Fan (2018).
\newblock Divide-and-{{Conquer}} for {{Debiased}} \$l\_1\$-norm {{Support
  Vector Machine}} in {{Ultra-high Dimensions}}.
\newblock {\em Journal of Machine Learning Research\/}~{\em 18\/}(182), 1--26.

\bibitem[\protect\citeauthoryear{Liu, Mao, and Zhang}{Liu
  et~al.}{2022}]{liu2022FastRobustSparsity}
Liu, W., X.~Mao, and X.~Zhang (2022).
\newblock Fast and {{Robust Sparsity Learning Over Networks}}: {{A
  Decentralized Surrogate Median Regression Approach}}.
\newblock {\em IEEE Transactions on Signal Processing\/}~{\em 70}, 797--809.

\bibitem[\protect\citeauthoryear{Lu, Zhu, and Lian}{Lu
  et~al.}{2023}]{lu2023SparseLowRankMatrix}
Lu, W., Z.~Zhu, and H.~Lian (2023).
\newblock Sparse and low-rank matrix quantile estimation with application to
  quadratic regression.
\newblock {\em Statistica Sinica\/}~{\em 33\/}(2), 945--959.

\bibitem[\protect\citeauthoryear{Luo and Tseng}{Luo and
  Tseng}{1992}]{luo1992LinearConvergenceDescent}
Luo, Z.-Q. and P.~Tseng (1992, March).
\newblock On the {{Linear Convergence}} of {{Descent Methods}} for {{Convex
  Essentially Smooth Minimization}}.
\newblock {\em SIAM Journal on Control and Optimization\/}~{\em 30\/}(2),
  408--425.

\bibitem[\protect\citeauthoryear{Mateos, Bazerque, and Giannakis}{Mateos
  et~al.}{2010}]{mateos2010DistributedSparseLinear}
Mateos, G., J.~A. Bazerque, and G.~B. Giannakis (2010, October).
\newblock Distributed {{Sparse Linear Regression}}.
\newblock {\em IEEE Transactions on Signal Processing\/}~{\em 58\/}(10),
  5262--5276.

\bibitem[\protect\citeauthoryear{Nedic and Ozdaglar}{Nedic and
  Ozdaglar}{2009}]{nedic2009DistributedSubgradientMethods}
Nedic, A. and A.~Ozdaglar (2009, January).
\newblock Distributed {{Subgradient Methods}} for {{Multi-Agent Optimization}}.
\newblock {\em IEEE Transactions on Automatic Control\/}~{\em 54\/}(1), 48--61.

\bibitem[\protect\citeauthoryear{Park, Kim, Myung, and Koo}{Park
  et~al.}{2012}]{park2012OraclePropertiesSCADpenalized}
Park, C., K.-R. Kim, R.~Myung, and J.-Y. Koo (2012).
\newblock Oracle properties of {{SCAD-penalized}} support vector machine.
\newblock {\em Journal of Statistical Planning and Inference\/}~{\em 142\/}(8),
  2257--2270.

\bibitem[\protect\citeauthoryear{Peng, Wang, and Wu}{Peng
  et~al.}{2016}]{peng2016ErrorBoundL1norm}
Peng, B., L.~Wang, and Y.~Wu (2016).
\newblock An error bound for l1-norm support vector machine coefficients in
  ultra-high dimension.
\newblock {\em Journal of Machine Learning Research\/}~{\em 17\/}(233), 1--26.

\bibitem[\protect\citeauthoryear{Platt}{Platt}{1998}]{platt1998SequentialMinimalOptimization}
Platt, J. (1998, April).
\newblock Sequential {{Minimal Optimization}}: {{A Fast Algorithm}} for
  {{Training Support Vector Machines}}.

\bibitem[\protect\citeauthoryear{Qiao and Chen}{Qiao and
  Chen}{2024}]{chen2022RobustFastLowRank}
Qiao, N. and C.~Chen (2024).
\newblock Fast and robust low-rank learning over networks: A decentralized
  matrix quantile regression approach.
\newblock {\em Journal of Computational and Graphical Statistics\/}~{\em
  33\/}(4), 1214–1223.

\bibitem[\protect\citeauthoryear{Qiao, Chen, and Zhu}{Qiao
  et~al.}{2025}]{chen2024DecentralizedDistributedEstimation}
Qiao, N., C.~Chen, and Z.~Zhu (2025).
\newblock Robust and efficient sparse learning over networks: a decentralized
  surrogate composite quantile regression approach.
\newblock {\em Statistics and Computing\/}~{\em 35\/}(1), 24.

\bibitem[\protect\citeauthoryear{Rosset and Zhu}{Rosset and
  Zhu}{2007}]{rosset2007PiecewiseLinearRegularized}
Rosset, S. and J.~Zhu (2007, July).
\newblock Piecewise linear regularized solution paths.
\newblock {\em The Annals of Statistics\/}~{\em 35\/}(3), 1012--1030.

\bibitem[\protect\citeauthoryear{Rubinstein}{Rubinstein}{1983}]{rubinstein1983SmoothedFunctionalsStochastic}
Rubinstein, R.~Y. (1983, February).
\newblock Smoothed {{Functionals}} in {{Stochastic Optimization}}.
\newblock {\em Mathematics of Operations Research\/}~{\em 8\/}(1), 26--33.

\bibitem[\protect\citeauthoryear{Sayed}{Sayed}{2014}]{sayed2014AdaptationLearningOptimization}
Sayed, A. (2014).
\newblock Adaptation, {{Learning}}, and {{Optimization}} over {{Networks}}.
\newblock {\em Foundations and Trends\textregistered{} in Machine
  Learning\/}~{\em 7\/}(4-5), 311--801.

\bibitem[\protect\citeauthoryear{Shi, Ling, Wu, and Yin}{Shi
  et~al.}{2015}]{shi2015ProximalGradientAlgorithm}
Shi, W., Q.~Ling, G.~Wu, and W.~Yin (2015).
\newblock A {{Proximal Gradient Algorithm}} for {{Decentralized Composite
  Optimization}}.
\newblock {\em IEEE Transactions on Signal Processing\/}~{\em 63\/}(22),
  6013--6023.

\bibitem[\protect\citeauthoryear{Tan, Wang, and Zhou}{Tan
  et~al.}{2022}]{tan2021HighDimensionalQuantileRegression}
Tan, K.~M., L.~Wang, and W.-X. Zhou (2022).
\newblock High-dimensional quantile regression: Convolution smoothing and
  concave regularization.
\newblock {\em Journal of the Royal Statistical Society Series B: Statistical
  Methodology\/}~{\em 84\/}(1), 205--233.

\bibitem[\protect\citeauthoryear{Tang, Zhou, and Song}{Tang
  et~al.}{2020}]{tang2020DistributedSimultaneousInference}
Tang, L., L.~Zhou, and P.~X.-K. Song (2020, March).
\newblock Distributed simultaneous inference in generalized linear models via
  confidence distribution.
\newblock {\em Journal of Multivariate Analysis\/}~{\em 176}, 104567.

\bibitem[\protect\citeauthoryear{Tseng}{Tseng}{2001}]{tseng2001ConvergenceBlockCoordinate}
Tseng, P. (2001, June).
\newblock Convergence of a {{Block Coordinate Descent Method}} for
  {{Nondifferentiable Minimization}}.
\newblock {\em Journal of Optimization Theory and Applications\/}~{\em
  109\/}(3), 475--494.

\bibitem[\protect\citeauthoryear{Vapnik}{Vapnik}{2000}]{vapnik2000NatureStatisticalLearning}
Vapnik, V.~N. (2000).
\newblock {\em The {{Nature}} of {{Statistical Learning Theory}}\/} (Second
  ed.).
\newblock {New York, NY}: {Springer New York}.

\bibitem[\protect\citeauthoryear{Wainwright}{Wainwright}{2019}]{wainwright2019HighDimensionalStatisticsNonAsymptotic}
Wainwright, M.~J. (2019, February).
\newblock {\em High-{{Dimensional Statistics}}: {{A Non-Asymptotic
  Viewpoint}}\/} (First ed.).
\newblock {Cambridge University Press}.

\bibitem[\protect\citeauthoryear{Wang, Zhou, Gu, and Zou}{Wang
  et~al.}{2022}]{wang2022DensityConvolutedSupportVector}
Wang, B., L.~Zhou, Y.~Gu, and H.~Zou (2022).
\newblock Density-{{Convoluted Support Vector Machines}} for {{High-Dimensional
  Classification}}.
\newblock {\em IEEE Transactions on Information Theory\/}~{\em 69\/}(4),
  2523--2536.

\bibitem[\protect\citeauthoryear{Wang and Li}{Wang and
  Li}{2018}]{wang2018DistributedQuantileRegression}
Wang, H. and C.~Li (2018, June).
\newblock Distributed {{Quantile Regression Over Sensor Networks}}.
\newblock {\em IEEE Transactions on Signal and Information Processing over
  Networks\/}~{\em 4\/}(2), 338--348.

\bibitem[\protect\citeauthoryear{Wang, Xia, and Li}{Wang
  et~al.}{2019}]{wang2019DistributedOnlineQuantile}
Wang, H., L.~Xia, and C.~Li (2019, April).
\newblock Distributed online quantile regression over networks with quantized
  communication.
\newblock {\em Signal Processing\/}~{\em 157}, 141--150.

\bibitem[\protect\citeauthoryear{Wang, Zhu, and Zou}{Wang
  et~al.}{2006}]{wang2006DoublyRegularizedSupport}
Wang, L., J.~Zhu, and H.~Zou (2006).
\newblock The doubly regularized support vector machine.
\newblock {\em Statistica Sinica\/}~{\em 16\/}(2), 589--615.

\bibitem[\protect\citeauthoryear{Wang, Zhu, and Zou}{Wang
  et~al.}{2008}]{wang2008HybridHuberizedSupport}
Wang, L., J.~Zhu, and H.~Zou (2008, February).
\newblock Hybrid huberized support vector machines for microarray
  classification and gene selection.
\newblock {\em Bioinformatics\/}~{\em 24\/}(3), 412--419.

\bibitem[\protect\citeauthoryear{Wang, Yang, Chen, and Liu}{Wang
  et~al.}{2019}]{wang2019DistributedInferenceLinear}
Wang, X., Z.~Yang, X.~Chen, and W.~Liu (2019).
\newblock Distributed inference for linear support vector machine.
\newblock {\em Journal of Machine Learning Research\/}~{\em 20}, 113:1--113:41.

\bibitem[\protect\citeauthoryear{Xu, Liu, and Lian}{Xu
  et~al.}{2024}]{xu2022DistributedEstimationSupporta}
Xu, W., J.~Liu, and H.~Lian (2024).
\newblock Distributed {{Estimation}} of {{Support Vector Machines}} for
  {{Matrix Data}}.
\newblock {\em IEEE Transactions on Neural Networks and Learning
  Systems\/}~{\em 35\/}(5), 6643--6653.

\bibitem[\protect\citeauthoryear{Yadav and Salapaka}{Yadav and
  Salapaka}{2007}]{yadav2007distributed}
Yadav, V. and M.~V. Salapaka (2007).
\newblock Distributed protocol for determining when averaging consensus is
  reached.
\newblock In {\em 45th Annual Allerton Conf}, pp.\  715--720.

\bibitem[\protect\citeauthoryear{Zhang}{Zhang}{2010}]{zhang2010NearlyUnbiasedVariable}
Zhang, C.-H. (2010, April).
\newblock Nearly unbiased variable selection under minimax concave penalty.
\newblock {\em The Annals of Statistics\/}~{\em 38\/}(2), 894--942.

\bibitem[\protect\citeauthoryear{Zhang, You, and Ba{\c s}ar}{Zhang
  et~al.}{2019}]{zhang2019DistributedDiscreteTimeOptimization}
Zhang, J., K.~You, and T.~Ba{\c s}ar (2019, June).
\newblock Distributed {{Discrete-Time Optimization}} in {{Multiagent Networks
  Using Only Sign}} of {{Relative State}}.
\newblock {\em IEEE Transactions on Automatic Control\/}~{\em 64\/}(6),
  2352--2367.

\bibitem[\protect\citeauthoryear{Zhang, Wu, Wang, and Li}{Zhang
  et~al.}{2016}]{zhang2016consistent}
Zhang, X., Y.~Wu, L.~Wang, and R.~Li (2016).
\newblock A consistent information criterion for support vector machines in
  diverging model spaces.
\newblock {\em Journal of Machine Learning Research\/}~{\em 17\/}(16), 1--26.

\bibitem[\protect\citeauthoryear{Zhou, She, and Song}{Zhou
  et~al.}{2023}]{zhou2024DistributedEmpiricalLikelihooda}
Zhou, L., X.~She, and P.~X.-K. Song (2023).
\newblock Distributed empirical likelihood approach to integrating unbalanced
  datasets.
\newblock {\em Statistica Sinica\/}~{\em 33\/}(3), 2209--2231.

\bibitem[\protect\citeauthoryear{Zhou and Shen}{Zhou and
  Shen}{2022}]{zhou2022CommunicationEfficientDistributedLearning}
Zhou, X. and H.~Shen (2022, January).
\newblock Communication-{{Efficient Distributed Learning}} for
  {{High-Dimensional Support Vector Machines}}.
\newblock {\em Mathematics\/}~{\em 10\/}(7), 1029.

\bibitem[\protect\citeauthoryear{Zhu, Rosset, Tibshirani, and Hastie}{Zhu
  et~al.}{2003}]{zhu20031normSupportVector}
Zhu, J., S.~Rosset, R.~Tibshirani, and T.~Hastie (2003).
\newblock 1-norm {{Support Vector Machines}}.
\newblock In {\em Advances in {{Neural Information Processing Systems}}},
  Volume~16. {MIT Press}.

\bibitem[\protect\citeauthoryear{Zou}{Zou}{2006}]{Zou2006TheAL}
Zou, H. (2006).
\newblock The adaptive lasso and its oracle properties.
\newblock {\em Journal of the American Statistical Association\/}~{\em
  101\/}(476), 1418--1429.

\bibitem[\protect\citeauthoryear{Zou and Hastie}{Zou and
  Hastie}{2005}]{zou2005RegularizationVariableSelection}
Zou, H. and T.~Hastie (2005, April).
\newblock Regularization and variable selection via the elastic net.
\newblock {\em Journal of the Royal Statistical Society: Series B (Statistical
  Methodology)\/}~{\em 67\/}(2), 301--320.

\bibitem[\protect\citeauthoryear{Zou and Li}{Zou and
  Li}{2008}]{zou2008OnestepSparseEstimates}
Zou, H. and R.~Li (2008, August).
\newblock One-step sparse estimates in nonconcave penalized likelihood models.
\newblock {\em The Annals of Statistics\/}~{\em 36\/}(4), 1509--1533.

\end{thebibliography}


\begin{thebibliography}{xx}

\harvarditem{Cai \harvardand\ Liu}{2011}{cai2011AdaptiveThresholdingSparse}
Cai, T. \harvardand\ Liu, W.  \harvardyearleft 2011\harvardyearright ,
  `Adaptive {{Thresholding}} for {{Sparse Covariance Matrix Estimation}}', {\em
  Journal of the American Statistical Association} {\bf 106}(494),~672--684.

\harvarditem[Cai et~al.]{Cai, Zhang \harvardand\
  Zhou}{2010}{cai2010OptimalRatesConvergence}
Cai, T.~T., Zhang, C.-H. \harvardand\ Zhou, H.~H.  \harvardyearleft
  2010\harvardyearright , `Optimal rates of convergence for covariance matrix
  estimation', {\em The Annals of Statistics} {\bf 38}(4),~2118--2144.

\harvarditem[Fan et~al.]{Fan, Liu, Sun \harvardand\
  Zhang}{2018}{fan2018ILAMMSparseLearning}
Fan, J., Liu, H., Sun, Q. \harvardand\ Zhang, T.  \harvardyearleft
  2018\harvardyearright , `I-{{LAMM}} for sparse learning: {{Simultaneous}}
  control of algorithmic complexity and statistical error', {\em The Annals of
  Statistics} {\bf 46}(2),~814--862.

\harvarditem[G{\"o}tze et~al.]{G{\"o}tze, Sambale \harvardand\
  Sinulis}{2021}{gotze2021ConcentrationInequalitiesPolynomials}
G{\"o}tze, F., Sambale, H. \harvardand\ Sinulis, A.  \harvardyearleft
  2021\harvardyearright , `Concentration inequalities for polynomials in
  {$\alpha$}-sub-exponential random variables', {\em Electronic Journal of
  Probability} {\bf 26}(none).

\harvarditem{Hong \harvardand\ Luo}{2017}{hong2017LinearConvergenceAlternating}
Hong, M. \harvardand\ Luo, Z.-Q.  \harvardyearleft 2017\harvardyearright , `On
  the linear convergence of the alternating direction method of multipliers',
  {\em Mathematical Programming} {\bf 162}(1),~165--199.

\harvarditem[Koo et~al.]{Koo, Lee, Kim \harvardand\
  Park}{2008}{koo2008BahadurRepresentationLinear}
Koo, J.-Y., Lee, Y., Kim, Y. \harvardand\ Park, C.  \harvardyearleft
  2008\harvardyearright , `A {{Bahadur Representation}} of the {{Linear Support
  Vector Machine}}.', {\em Journal of Machine Learning Research} {\bf
  9},~1343--1368.

\harvarditem{Ledoux \harvardand\
  Talagrand}{2011}{ledoux2011ProbabilityBanachSpaces}
Ledoux, M. \harvardand\ Talagrand, M.  \harvardyearleft 2011\harvardyearright ,
  {\em Probability in {{Banach}} Spaces: Isoperimetry and Processes}, Classics
  in Mathematics, {Springer}, {Berlin ; London}.

\harvarditem[Mateos et~al.]{Mateos, Bazerque \harvardand\
  Giannakis}{2010}{mateos2010DistributedSparseLinear}
Mateos, G., Bazerque, J.~A. \harvardand\ Giannakis, G.~B.  \harvardyearleft
  2010\harvardyearright , `Distributed {{Sparse Linear Regression}}', {\em IEEE
  Transactions on Signal Processing} {\bf 58}(10),~5262--5276.

\harvarditem[Sun et~al.]{Sun, Babu \harvardand\
  Palomar}{2017}{sun2017MajorizationMinimizationAlgorithmsSignal}
Sun, Y., Babu, P. \harvardand\ Palomar, D.~P.  \harvardyearleft
  2017\harvardyearright , `Majorization-{{Minimization Algorithms}} in {{Signal
  Processing}}, {{Communications}}, and {{Machine Learning}}', {\em IEEE
  Transactions on Signal Processing} {\bf 65}(3),~794--816.

\harvarditem{van~der Vaart \harvardand\
  Wellner}{2000}{vaart2000WeakConvergenceEmpirical}
van~der Vaart, A.~W. \harvardand\ Wellner, J.~A.  \harvardyearleft
  2000\harvardyearright , {\em Weak {{Convergence}} and {{Empirical
  Processes}}: {{With Applications}} to {{Statistics}}}, {Springer}, {New
  York}.

\harvarditem{Vershynin}{2018}{vershynin2018high}
Vershynin, R.  \harvardyearleft 2018\harvardyearright , {\em High-Dimensional
  Probability, An Introduction with Applications in Data Science}, {Cambridge
  University Press}, {Cambridge}.

\harvarditem{Wainwright}{2019}{wainwright2019HighDimensionalStatisticsNonAsymptotic}
Wainwright, M.~J.  \harvardyearleft 2019\harvardyearright , {\em
  High-{{Dimensional Statistics}}: {{A Non-Asymptotic Viewpoint}}}, first edn,
  {Cambridge University Press}.

\end{thebibliography}
\end{document}